\documentclass{article} % For LaTeX2e
\usepackage{nips13submit_e,times}
\usepackage{hyperref}
\usepackage{url}
\usepackage{epsfig}
\usepackage{verbatim}
\usepackage{amsmath,amsfonts}
\usepackage[mathscr]{eucal}
\usepackage{latexsym}
\usepackage{amscd}
\usepackage{amsthm}
\usepackage{graphics}
\usepackage{subfigure}
\usepackage[applemac]{inputenc}

\theoremstyle{remark}

\numberwithin{equation}{section}

\newcommand {\R}{\mathbb {R}}

\DeclareMathOperator*{\argmin}{arg\,min}

\makeatletter
\renewcommand*\env@matrix[1][*\c@MaxMatrixCols c]{%
  \hskip -\arraycolsep
  \let\@ifnextchar\new@ifnextchar
  \array{#1}}
\makeatother

\title{Spectral Networks and Deep Locally Connected Networks on Graphs}

\author{
Joan Bruna
\\
New York University\\
\texttt{bruna@cims.nyu.edu} \\
\And
Wojciech Zaremba \\
New York University\\
\texttt{woj.zaremba@gmail.com} \\
\AND
Arthur Szlam \\
The City College of New York\\
\texttt{aszlam@ccny.cuny.edu} \\
\And
Yann LeCun \\
New York University\\
\texttt{yann@cs.nyu.edu} \\
}

\nipsfinalcopy % Uncomment for camera-ready version

\begin{document}

\maketitle

\begin{abstract}
Convolutional Neural Networks are extremely efficient architectures
in image and audio recognition tasks, thanks to their ability to exploit
the local translational invariance of signal classes over their 
domain.
In this paper we consider possible generalizations of CNNs to signals
defined on more general domains without the action of a translation group. 
In particular, we propose two constructions, one based upon a hierarchical
clustering of the domain, and another based on the spectrum of the graph 
Laplacian. We show through experiments that for low-dimensional 
graphs it is possible to learn convolutional layers with a number of 
parameters independent of the input size,
resulting in efficient deep architectures. 

\end{abstract}

\section{Introduction}

Convolutional Neural Networks (CNNs) have been extremely succesful in machine learning problems where 
the coordinates of the underlying data representation have a grid structure (in 1, 2 and 3 dimensions), 
and the data to be studied in those coordinates has translational equivariance/invariance with respect 
to this grid. Speech \cite{deepSpeechReviewSPM2012}, images \cite{krizhevsky2012imagenet, lecun-01a, sermanet-icpr-12} or video \cite{taylor-eccv-10,le2011learning} are prominent examples that fall into this category.  

On a regular grid,
% for simplicity say a 1$d$ grid with $d=2^p$ points (so $Z_d$), 
a CNN is able to exploit several structures that play nicely together to greatly 
reduce the number of parameters in the system:
\begin{enumerate}
\item The translation structure, allowing the use of filters instead of generic linear maps and hence weight sharing.
\item The metric on the grid, allowing compactly supported filters, whose support is typically much 
smaller than the size of the input signals.
\item The multiscale dyadic clustering of the grid, allowing subsampling, implemented through stride convolutions and pooling.
\end{enumerate}

If there are $n$ input coordinates on a grid in $d$ dimensions, a fully connected layer with $m$ outputs requires $n\cdot m$ parameters, 
which in typical operating regimes amounts to a complexity of $O(n^2)$ parameters.
Using arbitrary filters instead of generic fully connected layers reduces the complexity 
to $O(n)$ parameters per feature map, as does using the metric structure by 
building a ``locally connected'' net \cite{DBLP:journals/corr/abs-1006-0448,Le10tiledconvolutional}. 
Using the two together gives $O(k \cdot S)$ parameters, where $k$ is the number 
of feature maps and $S$ is the support of the filters, and as a result 
the learning complexity is independent of $n$. 
  Finally, using the multiscale dyadic clustering allows 
each succesive layer to use a factor of $2^d$ less (spatial) coordinates per filter.

In many contexts, however, one may be faced with data defined over 
coordinates which lack some, or all, of the above 
geometrical properties. For instance, data defined on 3-D meshes, such as surface tension or temperature, 
measurements from a network of meteorological stations, or data coming from social networks or collaborative 
filtering, are all examples of structured inputs on which one cannot apply standard convolutional networks.
Another relevant example is the intermediate representation arising from deep neural networks. 
Although the spatial convolutional structure can be exploited at several layers, typical CNN architectures
do not assume any geometry in the ``feature" dimension, resulting in 4-D tensors which are only convolutional
along their spatial coordinates.

Graphs offer a natural framework to generalize the low-dimensional grid structure, and by extension 
the notion of convolution.
In this work, we will discuss constructions of deep neural networks on graphs other than regular grids. 
We propose two different constructions. In the first one, we show that one can extend
properties (2) and (3) to general graphs, and use them to define ``locally" connected and pooling layers, 
which require $O(n)$ parameters instead of $O(n^2)$. We term this the \emph{spatial} construction. 
The other construction, which we call \emph{spectral} construction, 
draws on the properties of convolutions in the Fourier domain. 
In $\mathbb{R}^d$, convolutions are linear operators diagonalised by the 
Fourier basis $\exp(i \omega \cdot t)$, $\omega,\,t\, \in \mathbb{R}^d$.
One may then extend convolutions to general graphs by finding the 
corresponding ``Fourier" basis.
This equivalence is given through the graph Laplacian, an operator 
which provides an harmonic analysis on the graphs \cite{belkin2001laplacian}.
The spectral construction needs at most $O(n)$ paramters per feature map, 
and also enables a construction where the number of parameters is independent
of the input dimension $n$.
These constructions allow efficient forward propagation and can be applied to 
datasets with very large number of coordinates.

%
%
%Decompositions with good localization in space and frequency. 
%
%Graphs with good local structure: in many applications, neighbors are meaningful, but
%far away coordinates are extremely noisy. 

\subsection{Contributions}

Our main contributions are summarized as follows:
\begin{itemize}
\item We show that from a weak geometric structure in the input domain it is possible to obtain
	 efficient architectures
	using $O(n)$ parameters, that we validate on low-dimensional graph 
	datasets.
\item We introduce a construction using $O(1)$ parameters which we empirically verify, and
	we discuss its connections with an harmonic analysis problem on graphs.
\end{itemize}

\section{Spatial Construction}

The most immediate generalisation of 
CNN to general graphs is to consider  multiscale, hierarchical, 
local receptive fields, as suggested in \cite{coates2011selecting}.
For that purpose, the grid will be replaced by a weighted graph 
$G=(\Omega,W)$, where $\Omega$ is 
a discrete set of size $m$ and $W$ is a $m\times m$  symmetric and nonnegative matrix.

\subsection{Locality via $W$}
The notion of locality can be generalized easily in the context of a graph.
Indeed, the weights in a graph determine a notion of locality. For example,
a straightforward way to define neighborhoods on $W$ is to set a threshold $\delta>0$ 
and take neighborhoods 
$$N_{\delta}(j)=\{i\in \Omega : W_{ij}>\delta\}~.$$  %discuss powers of W for bigger neighbor hoods here?
We can restrict attention to sparse ``filters'' with receptive fields given by these neighborhoods 
to get locally connected networks, thus 
reducing the number of parameters in a filter layer to $O(S \cdot n)$, where $S$ is 
the average neighborhood size.

\subsection{Multiresolution Analysis on Graphs}

CNNs reduce the size of the grid via pooling and subsampling layers.  
These layers are possible because of the natural multiscale clustering of the grid:  
they input all the feature maps over a cluster, and output a single feature for that cluster.  
On the grid, the dyadic clustering behaves nicely with respect 
to the metric and the Laplacian (and so with the translation structure). 
There is a large literature on forming multiscale clusterings on graphs, see for example 
\cite{Kushnir20061876,Luxburgspecttut,Dhillon:2007:WGC:1313055.1313291,Karypis95metis}.
% and although there is not any ``standard'' method, in many cases the methods described in those citations (or even simpler methods) can give good results.  
Finding multiscale clusterings that are provably guaranteed to behave well w.r.t. Laplacian on 
the graph is still an open area of research.  In this work we will use a naive agglomerative method.

\begin{figure}[h]
\centering
\includegraphics[scale=0.25]{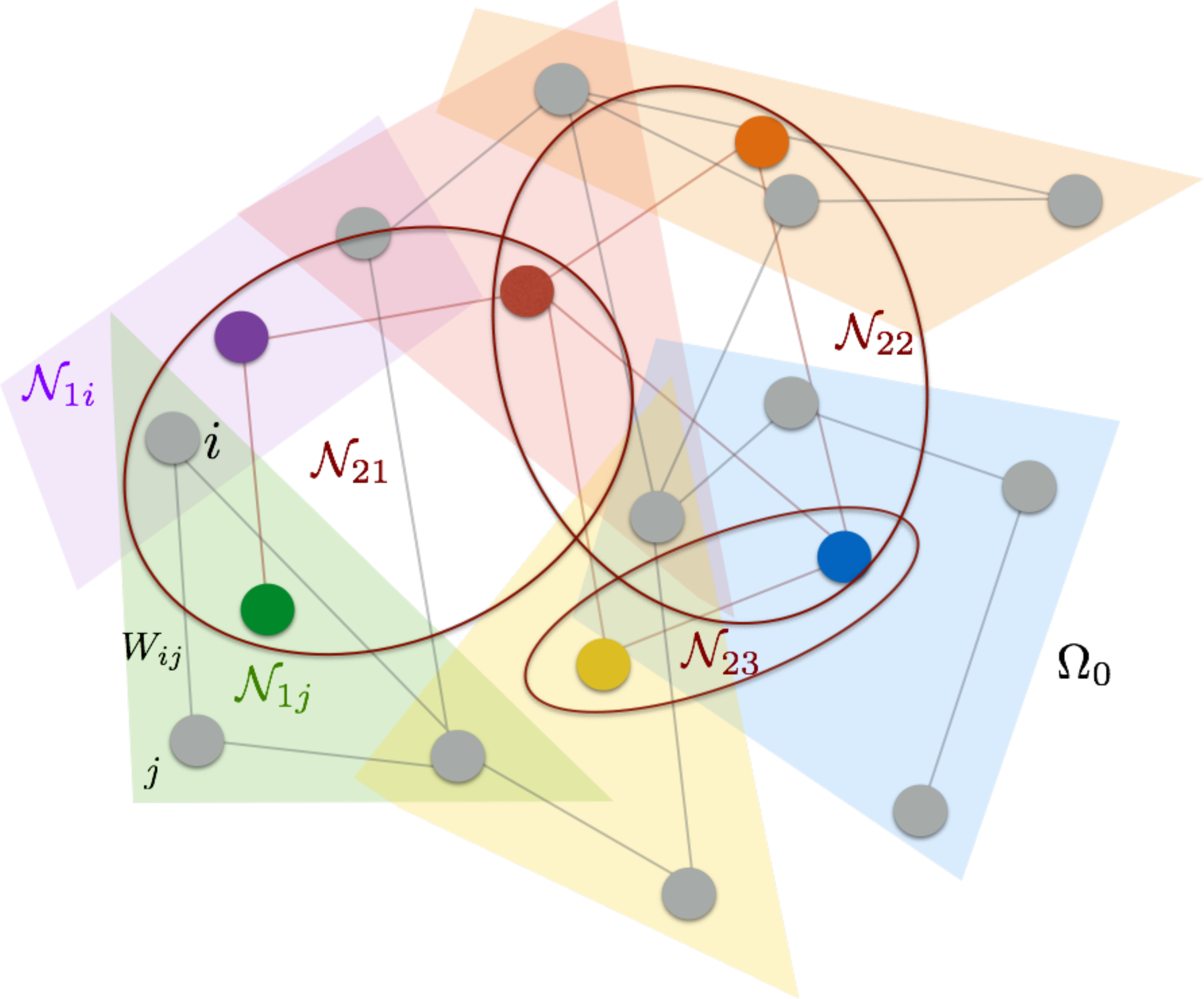}
\caption{Undirected Graph $G=(\Omega_0, W)$ with two levels of
clustering. The original points are drawn in gray.} 
\label{graph_diagram}
\end{figure}

Figure \ref{graph_diagram} illustrates a multiresolution clustering of a graph with 
the corresponding neighborhoods.

\subsection{Deep Locally Connected Networks}
\label{spatial_sec}

%Instead of generalizing a CNN by finding spectral multipliers, one can work entirely in space.  
The spatial construction starts with a 
 multiscale clustering of the graph, 
 similarly as in \cite{coates2011selecting}
We consider $K$ scales. We set $\Omega_0=\Omega$, 
and for each $k=1\dots K$, 
we define $\Omega_k$, a partition of $\Omega_{k-1}$ 
into $d_k$ clusters; and a collection of neighborhoods 
around each element of $\Omega_{k-1}$:
$$\mathcal{N}_k=\{ \mathcal{N}_{k,i}\,;\; i=1\dots d_{k-1} \}~.$$
 
 %$\Omega_k$ and neighborhoods at each scale.
 %  
%given by the rows of the matrices $W_k$ (the rows and columns of $W_k$ are indexed by 
%the clusters in $\Omega_{k-1}$, where $\Omega_0=\Omega$.
With these in hand, we can now define the $k$-th layer of the network.
We assume without loss of generality that the input signal is a real 
signal defined in $\Omega_0$, and we denote by $f_k$ the number of 
``filters" created at each layer $k$.
Each layer of the network will transform a 
 $f_{k-1}$-dimensional signal indexed by $\Omega_{k-1}$ into 
a $f_k$-dimensional signal indexed by $\Omega_k$, thus trading-off 
spatial resolution with newly created feature coordinates.
%If $f_k$ denotes the number of ``filters" created at layer $k$ and 
% this layer, 

\begin{figure}[h]
\centering
\includegraphics[scale=0.4]{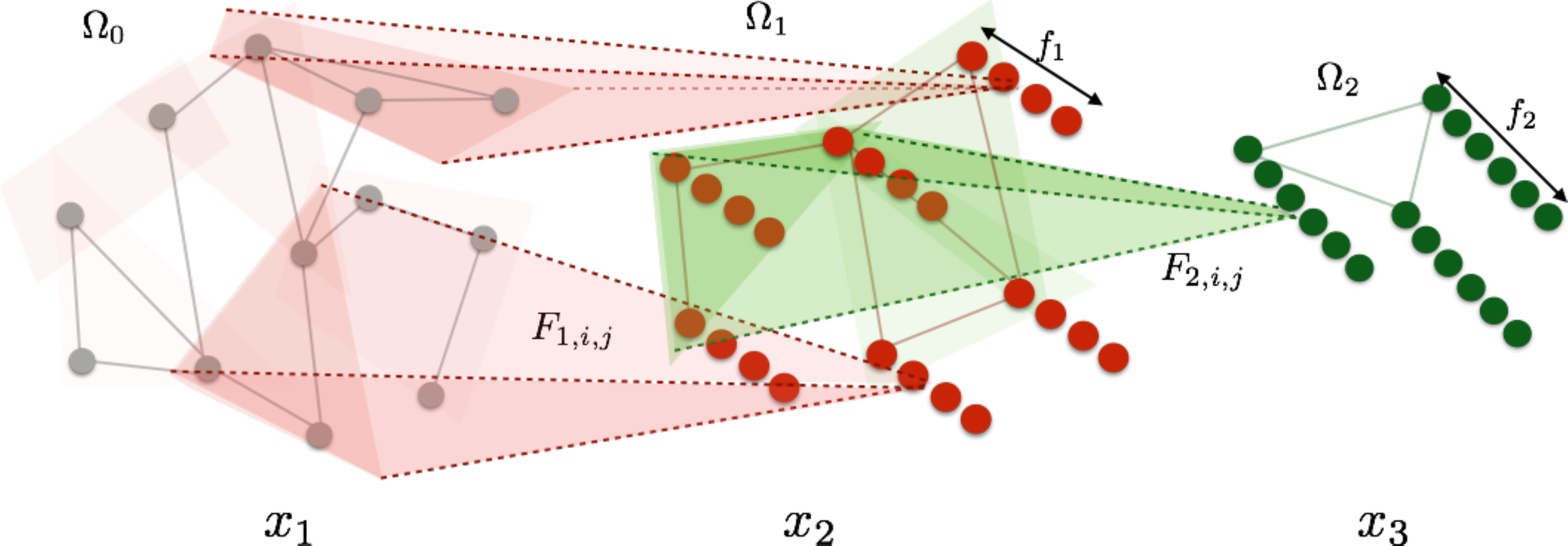}
\caption{Spatial Construction as described by (\ref{spatial_construction_eq}), with $K=2$. 
For illustration purposes, the pooling operation is assimilated with the filtering stage. 
Each layer of the transformation loses spatial resolution but increases the number of filters. } 
\label{spatial_construction_fig}
\end{figure}

More formally, if $x_k= ( x_{k,i} \, \, ; \, i=1\dots f_{k-1} )$ is the $d_{k-1}\times f_{k-1}$ is the 
input to layer $k$, its the output $x_{k+1}$ is defined as 
\begin{equation} 
\label{spatial_construction_eq}
x_{k+1, j} =   L_k h \left( \sum_{i=1}^{f_{k-1}} F_{k,i,j} x_{k,i} \right)~~(j=1\dots f_k)~,
\end{equation}
%where $h:\R\rightarrow \R$ is a point-wise nonlinearity,
%$x_k$ is the $d_{k-1}\times f_{k-1}$ input vector, where $d_k$ is the number of clusters at 
%level $k$, $f_k$ is the number of ``filters" at level $k$, and $x_k$ is the filter responses vertically concatenated.  Furthermore, 
%\begin{equation} 
%F_k=
%\begin{pmatrix}
%F_{k,1,1}    & ... &F_{k,1,f_{k-1}}\\
%\vdots       &     &\vdots     \\
%F_{k,f_k,1}    & ...&F_{k,f_k,f_{k-1}}\\
%\end{pmatrix},
%\end{equation}
where $F_{k,i,j}$ is a $d_{k-1}\times d_{k-1}$ sparse matrix with nonzero entries in the 
locations given by $\mathcal{N}_k$, 
%same locations as 
%$w_{k-1}$.   $h$ is again a componentwise nonlinearity, 
and $L_k$ outputs the result of a pooling operation over each cluster in $\Omega_k$.
This construcion is illustrated in Figure \ref{spatial_construction_fig}.

In the current code, to build $\Omega_k$ and $\mathcal{N}_k$ we use the following construction: 
\begin{eqnarray*}
W_0 & = & W \\
A_k(i,j)&=&\sum_{s\in \Omega_k(i)} \sum_{t\in \Omega_k(j)} W_{k-1}(s,t)~,~(k \leq K)\\
W_k &=&  \text{rownormalize}(A_k) ~,~(k \leq K) \\
\mathcal{N}_k &=& \mbox{supp}(W_k) ~.~(k \leq K)
\end{eqnarray*}
and $\Omega_k$ is found as an $\epsilon$ covering for $W_k$
 \footnote{An $\epsilon$-covering of a set $\Omega$ using a similarity kernel $K$ 
 is a partition $\mathcal{P}=\{\mathcal{P}_1,\dots,\mathcal{P}_n\}$ 
 such that $\sup_n \sup_{x,x'  \in \mathcal{P}_n} K(x,x') \geq \epsilon$.}.
This is just one amongst many strategies to perform hierarchicial 
agglomerative clustering. For a larger account of the problem, we refer the 
reader to \cite{hastie_09}.

If $S_k$ is the average support of the neighborhoods $\mathcal{N}_k$,
we verify from (\ref{spatial_construction_eq}) that the number of parameters 
to learn at layer $k$ is 
$$O(S_k \cdot |\Omega_k| \cdot f_k \cdot f_{k-1})=O(n)~.$$ 
In practice, we have $S_k \cdot |\Omega_k|  \approx \alpha \cdot | \Omega_{k-1}|$, 
where $\alpha$ is the oversampling factor, typically $\alpha \in (1,4)$. 

The spatial construction might appear naïve, but it has the advantage 
that it requires relatively weak regularity assumptions on the graph. 
Graphs having low intrinsic dimension have localized neighborhoods, 
even if no nice global embedding exists.
 However, under this construction there is 
 no easy way to induce weight sharing across different locations of the graph.
One possible option is to consider a global embedding of the graph into a low dimensional space, 
which is rare in practice for high-dimensional data.

\section{Spectral Construction}

The global structure of the graph can be exploited 
with the spectrum of its graph-Laplacian to 
generalize the convolution operator.

\subsection{Harmonic Analysis on Weighted Graphs}

   The combinatorial Laplacian $L=D-W$ or graph Laplacian 
   $\mathcal{L}=I-D^{-1/2}WD^{-1/2}$ are generalizations of the Laplacian on the grid; 
   and frequency and smoothness relative to $W$ are interrelated through these operators 
   \cite{Chung:1997,Luxburgspecttut}.   For simplicity, here we use the combinatorial Laplacian.  
   If $x$ is an $m$-dimensional vector, 
   a natural definition of the smoothness functional $||\nabla x||_W^2$ at a node $i$ is 
\[\|\nabla x\|_W^2(i)=\sum_j W_{ij}[x(i)-x(j)]^2,\]
and 
\begin{equation}
\label{smoothlaplacian}
\|\nabla x\|_W^2=\sum_i\sum_j W_{ij}[x(i)-x(j)]^2~,
\end{equation}

With this definition, the smoothest vector is a constant:  
\[v_0=\argmin_{x\in \R^m \,\, \|x\|=1} \|\nabla x\|_W^2=(1/\sqrt{m}) 1_m.\]  

Each succesive \[v_i=\argmin_{x\in \R^m \,\, \|x\|=1 \,\, x\perp \{v_0,...,v_{i-1}\}} \|\nabla x\|_W^2\] 
is an eigenvector of $L$, and the eigenvalues $\lambda_i$ allow the smoothness 
of a vector $x$ to be read off from the coefficients of $x$ 
in $[v_0,...v_{m-1}]$, equivalently as the Fourier coefficients of a 
signal defined in a grid.
Thus, just an in the case of the grid, where the eigenvectors of the Laplacian are the 
Fourier vectors,  diagonal operators on the spectrum of the Laplacian modulate the smoothness of their operands.  
Moreover, using these diagonal operators reduces the number of parameters of a filter from $m^2$ to $m$.  
%If smoothness on the graph (as defined by $||\nabla f||_W^2$) is reasonable

These three structures above are all tied together through the Laplacian 
operator on the $d$-dimensional grid $\Delta x = \sum_{i=1}^d \frac{\partial^2 x}{\partial u_i^2}$ :
\begin{enumerate}
\item Filters are multipliers on the eigenvalues of the Laplacian $\Delta$.
\item Functions that are smooth relative to the grid metric have coefficients with quick decay in the basis of eigenvectors of $\Delta$.
\item The eigenvectors of the subsampled Laplacian are the low frequency eigenvectors of $\Delta$.
\end{enumerate}

\subsection{Extending Convolutions via the Laplacian Spectrum}
\label{spectral_sec}

As in section \ref{spatial_sec}, 
let $W$ be a  weighted graph with index set denoted by $\Omega$, and let $V$
 be the eigenvectors of the graph Laplacian $L$, ordered by eigenvalue.
Given a weighted graph, we can try to generalize a convolutional net by operating on the spectrum of the 
weights, given by the eigenvectors of its graph Laplacian.

For simplicity, let us first describe a construction where each layer $k=1\dots K$ 
transforms an input vector $x_k$ of size $|\Omega| \times f_{k-1}$ into an output $x_{k+1}$ 
of dimensions $|\Omega| \times f_{k}$, that is, without spatial subsampling:
\begin{equation} 
\label{spectral_construction_eq}
x_{k+1, j} =   h \left( V \sum_{i=1}^{f_{k-1}} F_{k,i,j} V^T x_{k,i} \right)~~(j=1\dots f_k)~,
\end{equation}
where $F_{k,i,j}$ is a diagonal matrix and, as before, $h$ is a real valued nonlinearity.

Often, only the first $d$ eigenvectors of the Laplacian are useful in practice, which  
carry the smooth geometry of the graph. The cutoff frequency $d$ depends upon 
the intrinsic regularity of the graph and also the sample size.  
In that case, we can replace in (\ref{spectral_construction_eq}) $V$ by
$V_d$, obtained by keeping the first $d$ columns of $V$.

If the graph has an underlying group invariance this construction can discover it; the best example 
being the standard CNN; see \ref{sec:rediscover}. 
However,  in many cases the graph does not 
have a group structure, or the group structure does not commute with the Laplacian, 
and so we cannot think of each filter as passing a template across $\Omega$ and recording the correlation 
of the template with that location. 
$\Omega$ may not be homogenous in a way that allows this to make sense, as we shall 
see in the example from Section \ref{subsmnist}.

%Number of parameters
Assuming only $d$ eigenvectors of the Laplacian are kept, 
equation (\ref{spectral_construction_eq}) shows that each layer requires
$f_{k-1} \cdot f_k \cdot d = O(|\Omega|)$ paramters to train. 
We shall see in section \ref{smoothspect} how the global and local 
regularity of the graph can be combined to produce layers with $O(1)$ 
parameters, i.e. such that the number of learnable parameters does not depend upon 
the size of the input.

This construction can suffer from the fact that most graphs have meaningful eigenvectors 
only for the very top of the spectrum. 
Even when the individual high frequency eigenvectors are not 
meaningful, a cohort of high frequency eigenvectors may contain meaningful information.  However 
this construction may not be able to access this information because it is nearly diagonal at the highest frequencies.   
 
Finally,  it is not obvious how to do either the forwardprop or the backprop efficiently while applying the nonlinearity 
on the space side, as we have to make the expensive multiplications by $V$ and $V^T$; and it is not obvious 
how to do standard nonlinearities on the spectral side. However, see \ref{sec:multigrid}.  
 
%Why is the previous spectral construction not always good? 
%Stability with respect to noise / sampling. Only the regular part of the spectrum is in practice stable. 
%High frequencies are important for recognition and are not efficiently captured with 
%the Laplacian eigenmaps.
 
 \subsection{Rediscovering standard CNN's}
 \label{sec:rediscover}

A simple, and in some sense universal, choice of weight matrix in this construction 
is the covariance of the data.  
Let $X=(x_k)_k$ be the input data distribution, with $x_k \in \mathbb{R}^n$.
If each coordinate $j=1\dots n$ has the same variance,  
$$\sigma^2_j = E\left( | x(j) -  E(x(j))|^2\right)~,$$
then diagonal operators on the Laplacian simply scale the principal components of $X$. 
While this may seem trivial, it is well known that the principal components of the 
set of images of a fixed size are (experimentally) correspond to 
the Discrete Cosine Transform basis, 
organized by frequency. This can be explained by noticing that images 
are translation invariant, and hence the covariance operator 
$$\Sigma(j,j) = E\left( (x(j) -  E(x(j)))(x(j') -  E(x(j')))\right) $$
satisfies $\Sigma(j,j') = \Sigma(j-j')$, hence it is diagonalized by the Fourier basis. 
Moreover, it is well known that 
natural images exhibit a power spectrum $E(|\widehat{x}(\xi)|^2) \sim \xi^{-2} $, 
since nearby pixels are more correlated than far away pixels. 
It results that principal components of the covariance are essentially ordered from
low to high frequencies, which is consistent with the standard group structure of the 
Fourier basis.

The upshot is that, when applied to natural images, the construction in \ref{spectral_sec}  
using the covariance as the similarity kernel recovers a standard convolutional network,
 without any prior knowledge. Indeed, the linear operators $V F_{i,j} V^T$ from Eq (\ref{spectral_construction_eq}) 
 are by the previous argument diagonal in the Fourier basis, hence translation invariant, 
 hence ``classic" convolutions. Moreover, Section \ref{sec:multigrid} explains how spatial 
 subsampling can also be obtained via dropping the last part of the spectrum of the Laplacian, 
 leading to max-pooling, and ultimately to deep convolutonal networks.

\subsection{$O(1)$ construction with smooth spectral multipliers}
\label{smoothspect}

 In the standard grid, we do not need a parameter for each 
 Fourier function because the filters 
 are compactly supported in space, but in (\ref{spectral_construction_eq}), 
 each filter requires one parameter for each eigenvector on which it acts. 
 Even if the filters were compactly supported in space in this construction, 
 we still would not get less than $O(n)$ parameters per filter because the spatial 
 response would be different at each location.  
 
 One possibility for getting around this is to generalize the duality of the grid.  
 On the Euclidian grid, the decay of a function in the spatial domain is translated 
 into smoothness in the Fourier domain, and viceversa. 
It results that a funtion $x$ which is spatially localized 
  has a smooth frequency response $\hat{x} = V^T x$. 
In that case, the eigenvectors of the Laplacian can be thought of as being arranged on a grid 
isomorphic to the original spatial grid.

This suggests that, in order to learn a layer in which features will be not only 
shared across locations but also well localized in the original domain,
one can learn spectral multipliers which are smooth. 
Smoothness can be prescribed by learning only a subsampled set
of frequency multipliers and using an interpolation kernel 
to obtain the rest, such as cubic splines.
However, the notion of smoothness requires a geometry in the domain of spectral coordinates, 
which can be obtained by defining a dual graph $\widetilde{W}$ as shown by (\ref{smoothlaplacian}).
As previously discussed, on regular grids this geometry is 
given by the notion of frequency, but this cannot be 
directly generalized to other graphs. 

A particularly simple and navie choice consists in choosing a 
$1$-dimensional arrangement, obtained by ordering the eigenvectors according
to their eigenvalues. In this setting, 
the diagonal of each filter $F_{k,i,j}$ (of size at most $|\Omega|)$ is parametrized by 
$$\mbox{diag}(F_{k,i,j}) = \mathcal{K} \, \alpha_{k,i,j} ~,$$
where $\mathcal{K}$ is a $d \times q_k$ 
fixed cubic spline kernel and 
$\alpha_{k,i,j}$ are the $q_k$ spline coefficients. 
If one seeks to have filters with constant spatial support (ie, whose support is independent of the input size $|\Omega|$), 
it follows that one can choose a sampling step $\alpha \sim |\Omega|$ in the spectral domain, which 
results in a constant number $q_k \sim |\Omega| \cdot \alpha^{-1} = O(1)$ of coefficients $\alpha_{k,i,j}$ per filter.

Although results from section \ref{secexperiments} seem to indicate that the 1-D arrangement 
given by the spectrum of the Laplacian is efficient at creating spatially localized filters, a fundamental question
is how to define a dual graph capturing the geometry of spectral coordinates. 
A possible algorithmic stategy is to consider an input distribution $X=(x_k)_k$ consisting on spatially localized 
signals and to construct a dual graph $\widehat{W}$ by measuring the similarity of 
in the spectral domain: $\widehat{X}= V^T X$. 
The similarity could be measured for instance
with $E( ( |\hat{x}| -E(|\hat{x})|))^T( |\hat{x}| -E(|\hat{x}|))$.

%\subsubsection{Another $O(1)$ construction via local embeddings and kernel ridge regression}

\section{Relationship with previous work}

There is a large literature on building wavelets on graphs, see for example 
\cite{deep_wavelets,conf/icml/GavishNC10,CMDiffusionWavelets,conf/infocom/CrovellaK03,irrsub99}.  
A wavelet basis on a grid, in the language of neural networks, is a linear autoencoder with certain 
provable regularity properties (in particular, when encoding various classes of smooth functions, sparsity is guaranteed).  
The forward propagation in a classical wavelet transform strongly resembles the forward propagation in a neural 
network, except that there is only one filter map at each layer (and it is usually the same filter at each layer), and 
the output of each layer is kept, rather than just the output of the final layer.   
Classically, the filter is not learned, but constructed to facilitate the regularity proofs.  

In the graph case, the goal is the same; except that the smoothness on the grid is replaced by smoothness 
on the graph.  As in the classical case, most works have tried to construct the wavelets explicitly (that is, without learning), 
based on the graph, so that the corresponding autencoder has the correct sparsity properties.   
In this work, and the recent work \cite{deep_wavelets}, the ``filters'' are constrained by construction to 
have some of the regularity properties of wavelets, but are also trained so that they are appropriate for a 
task separate from (but perhaps related to) the smoothness on the graph.  Whereas \cite{deep_wavelets} 
still builds a (sparse) linear autoencoder that keeps the basic wavelet transform setup, this work focuses 
on nonlinear constructions; and in particular, tries to build analogues of CNN's.

Another line of work which is rellevant to the present work is that of discovering 
grid topologies from data. In \cite{le2007learning}, the authors empirically confirm the 
statements of Section \ref{sec:rediscover}, by showing that one can recover the 2-D 
grid structure via second order statistics. In \cite{coates2011selecting, jia2012beyond} 
the authors estimate similarities between features to construct locally connected networks.

\subsection{Multigrid}
\label{sec:multigrid}
We could improve both constructions, and to some extent unify them, with  a multiscale 
clustering of the graph that plays nicely with the Laplacian.  As mentioned before, 
in the case of the grid, the standard dyadic cubes have the property that subsampling the 
Fourier functions on the grid to a coarser grid is the same as finding the Fourier functions on the coarser grid.  
This property would eliminate the annoying necessity of mapping  the spectral construction 
to the finest grid at each layer to do the nonlinearity; and would allow us to interpret (via interpolation) 
the local filters at deeper layers in the spatial construction  to be low frequency.  

This kind of clustering is the underpinning of the multigrid method for solving discretized PDE's (and linear systems in general) 
\cite{Trottenberg:2000:MUL:374106}.   There have been several papers extending the multigrid 
method, and in particular, the multiscale clustering(s) associated to the multigrid method, in settings 
more general than regular grids, see for example \cite{Kushnir20061876,5184845} for situations as 
in this paper, and see \cite{Trottenberg:2000:MUL:374106} for the algebraic multigrid method in general. 
In this work, for simplicity, we use a naive multiscale clustering on the space side construction that is 
not guaranteed to respect the original graph's Laplacian, and no explicit spatial clustering in the spectral construction.

\section{Numerical Experiments}
\label{secexperiments}

The previous constructions are tested on two variations of the MNIST data set.  
In the first, we subsample the normal $28\times28$ grid to get $400$ coordinates. 
These coordinates still have a $2$-D structure, but it is not possible to use standard convolutions. 
We then make a dataset by placing $d=4096$ points on the 
$3$-D unit sphere  and project random MNIST images onto 
this set of points, as described in Section \ref{sphereMNISTsect}.

In all the experiments, we use Rectified Linear Units as nonlinearities and max-pooling.
We train the models with cross-entropy loss, using
 a fixed learning rate of $0.1$ with momentum $0.9$. 

\subsection{Subsampled MNIST}
\label{subsmnist}

\begin{figure}
\centering
\subfigure[ ]{
\includegraphics[scale=0.4]{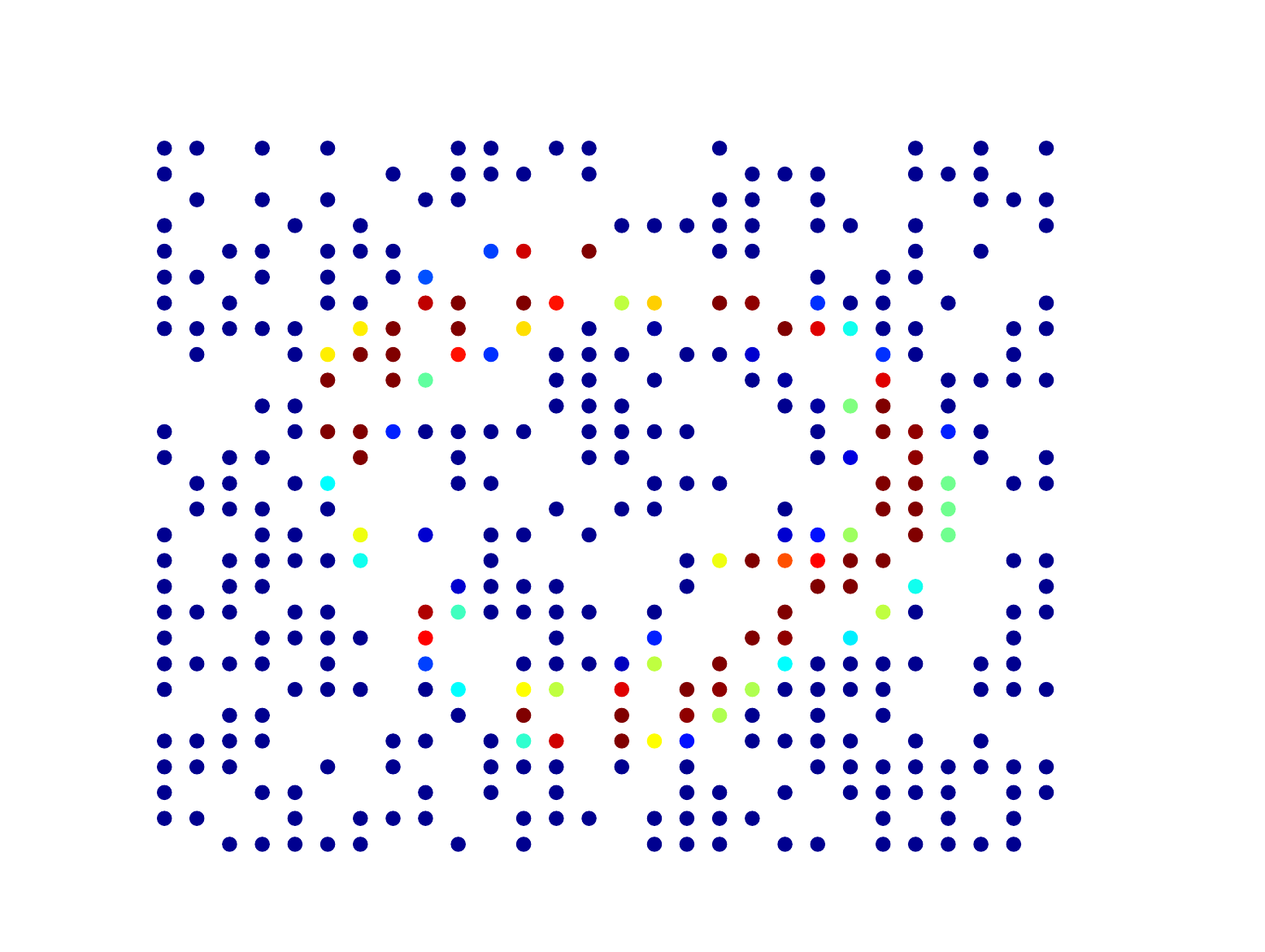}}
\subfigure[ ]{
\includegraphics[scale=0.4]{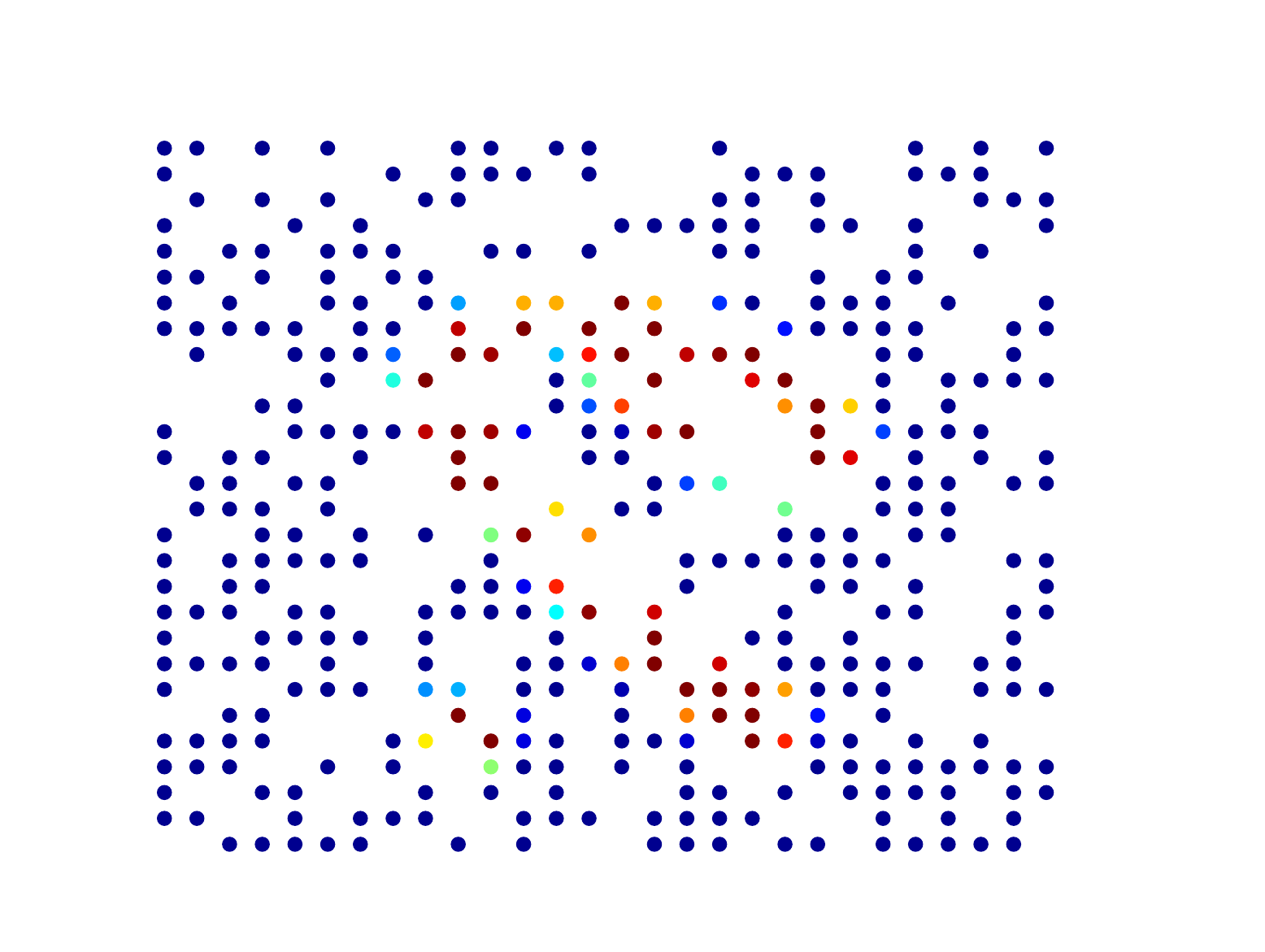}} 
\caption{Subsampled MNIST examples.} 
\label{mnistgrid_ex}
\end{figure}

\begin{figure}
\centering
\subfigure[ ]{
\includegraphics[scale=0.4]{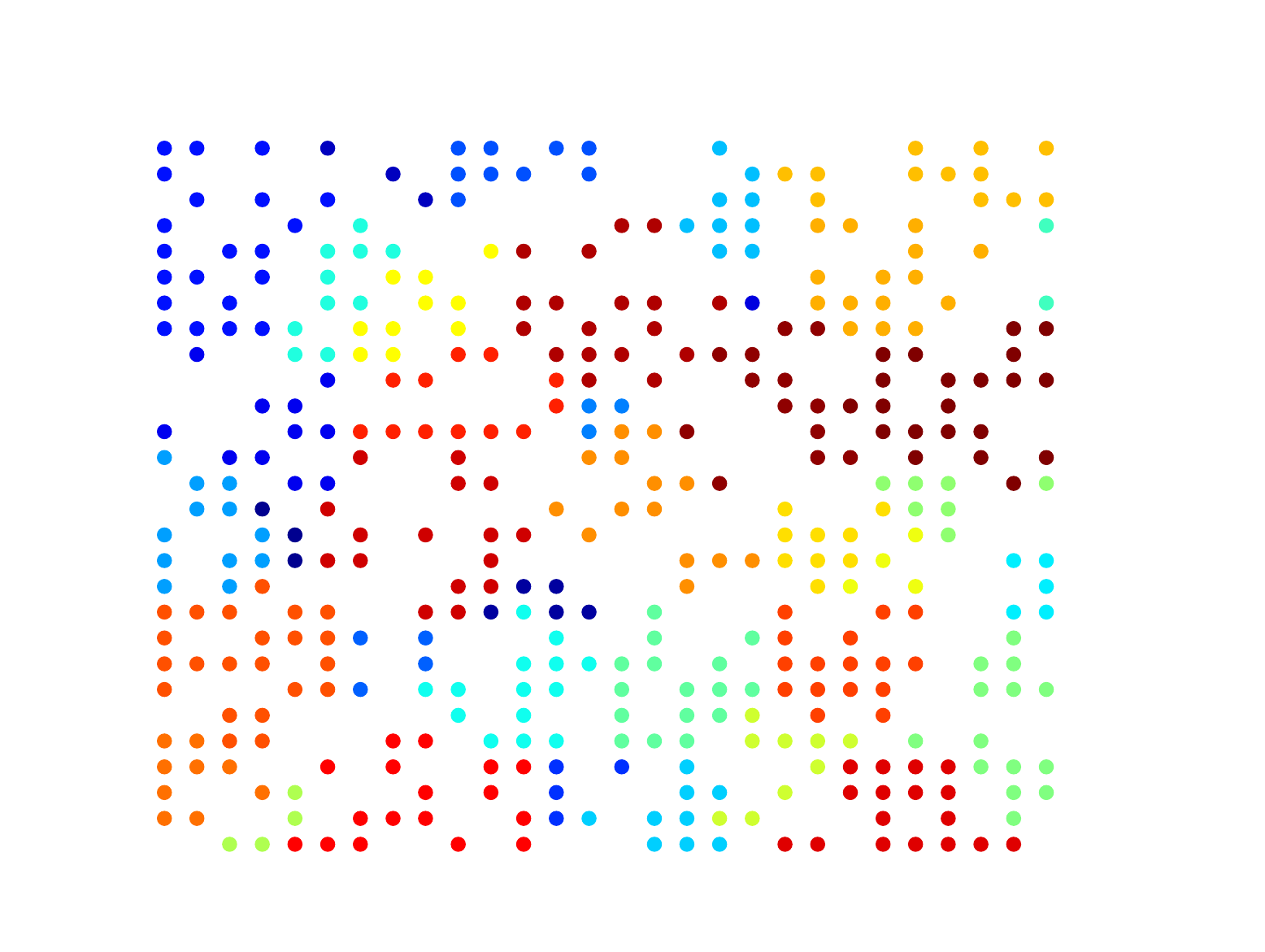}}
\subfigure[ ]{
\includegraphics[scale=0.4]{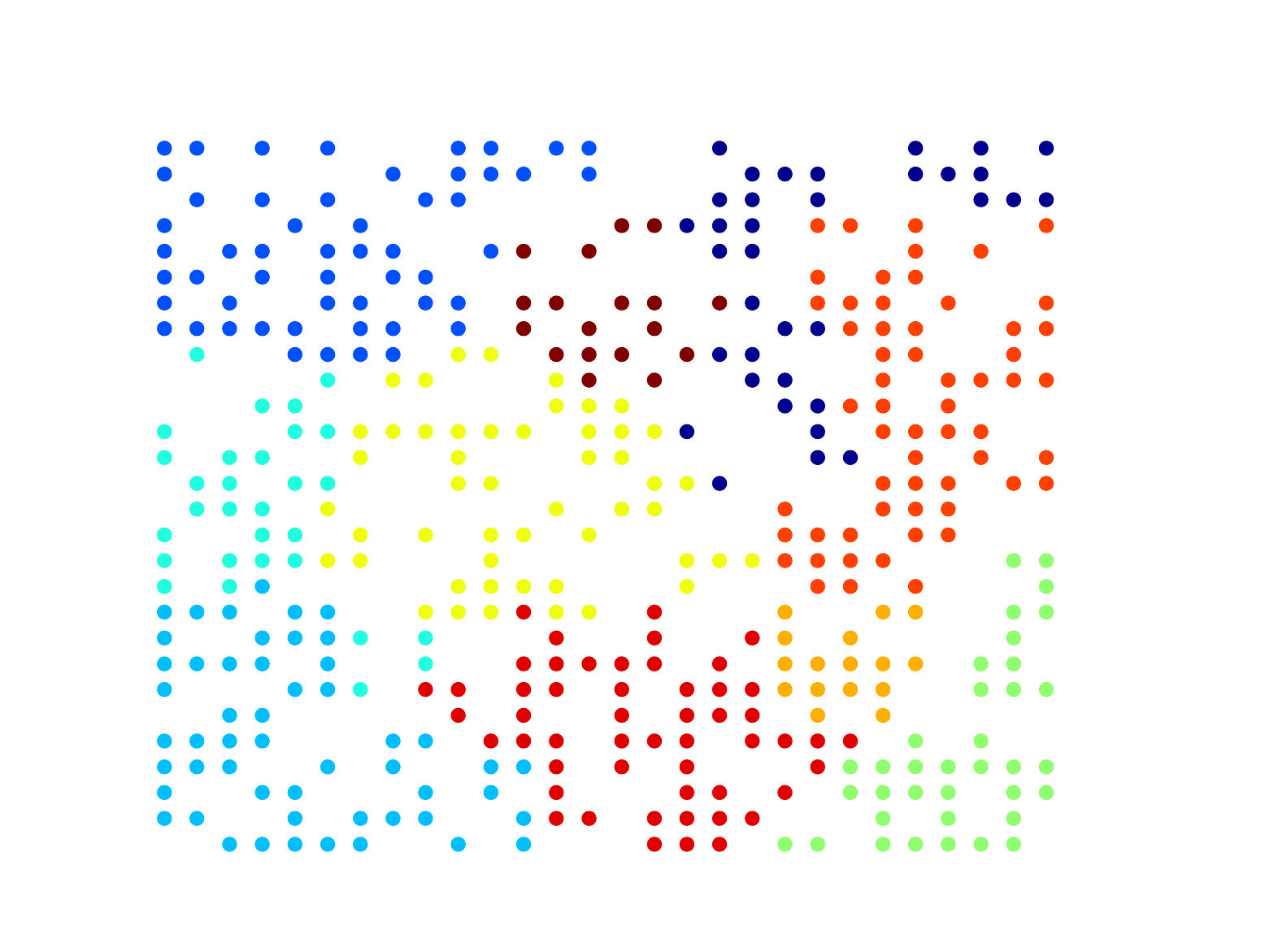}} 
\caption{Clusters obtained with 
the agglomerative clustering. 
(a) Clusters corresponding to the finest scale $k=1$, (b) clusters for $k=3$ .} 
\label{mnistgrid_space}
\end{figure}

\begin{figure}
\centering
\subfigure[ ]{
\includegraphics[scale=0.4]{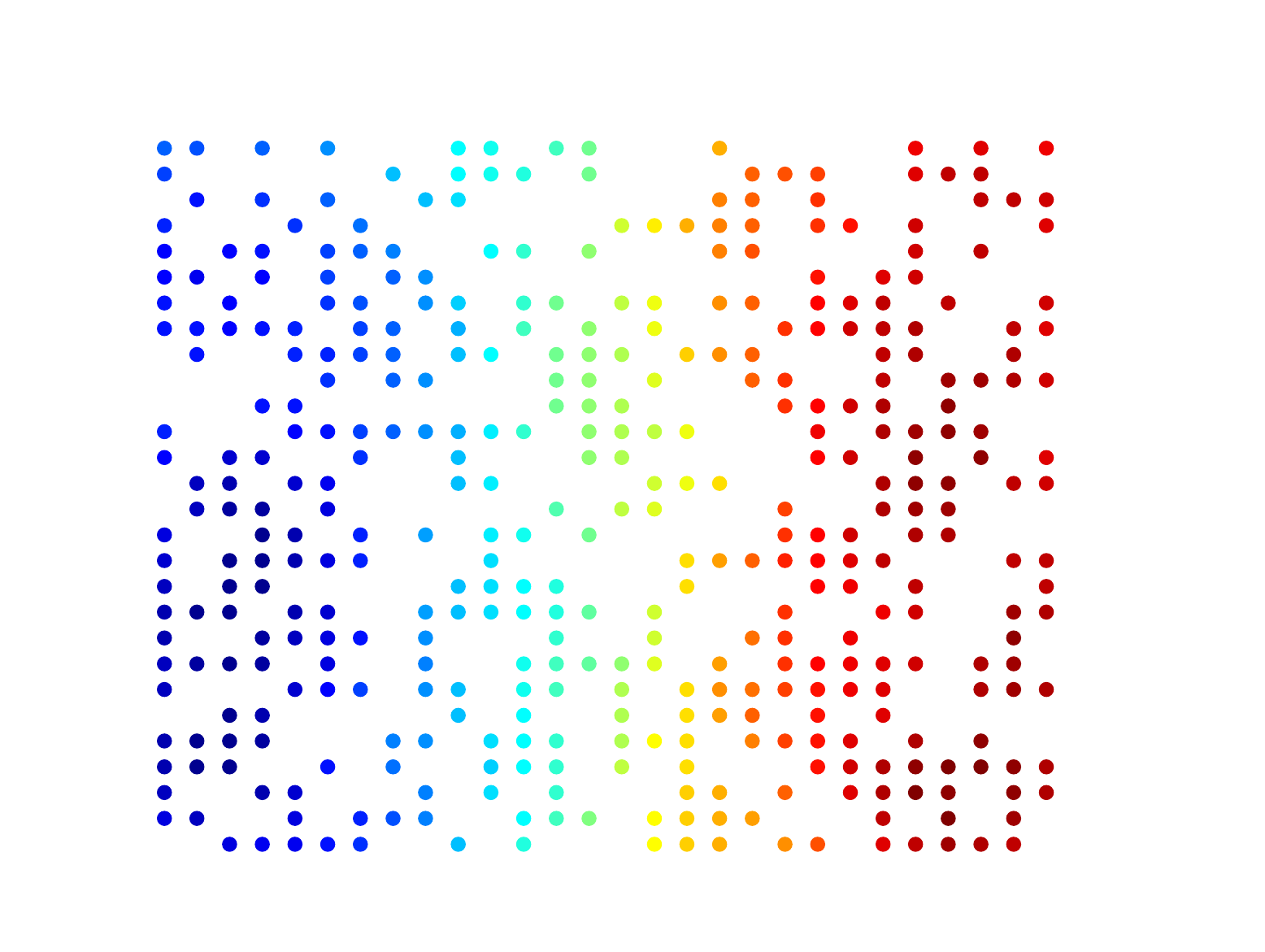}}
\subfigure[ ]{
\includegraphics[scale=0.4]{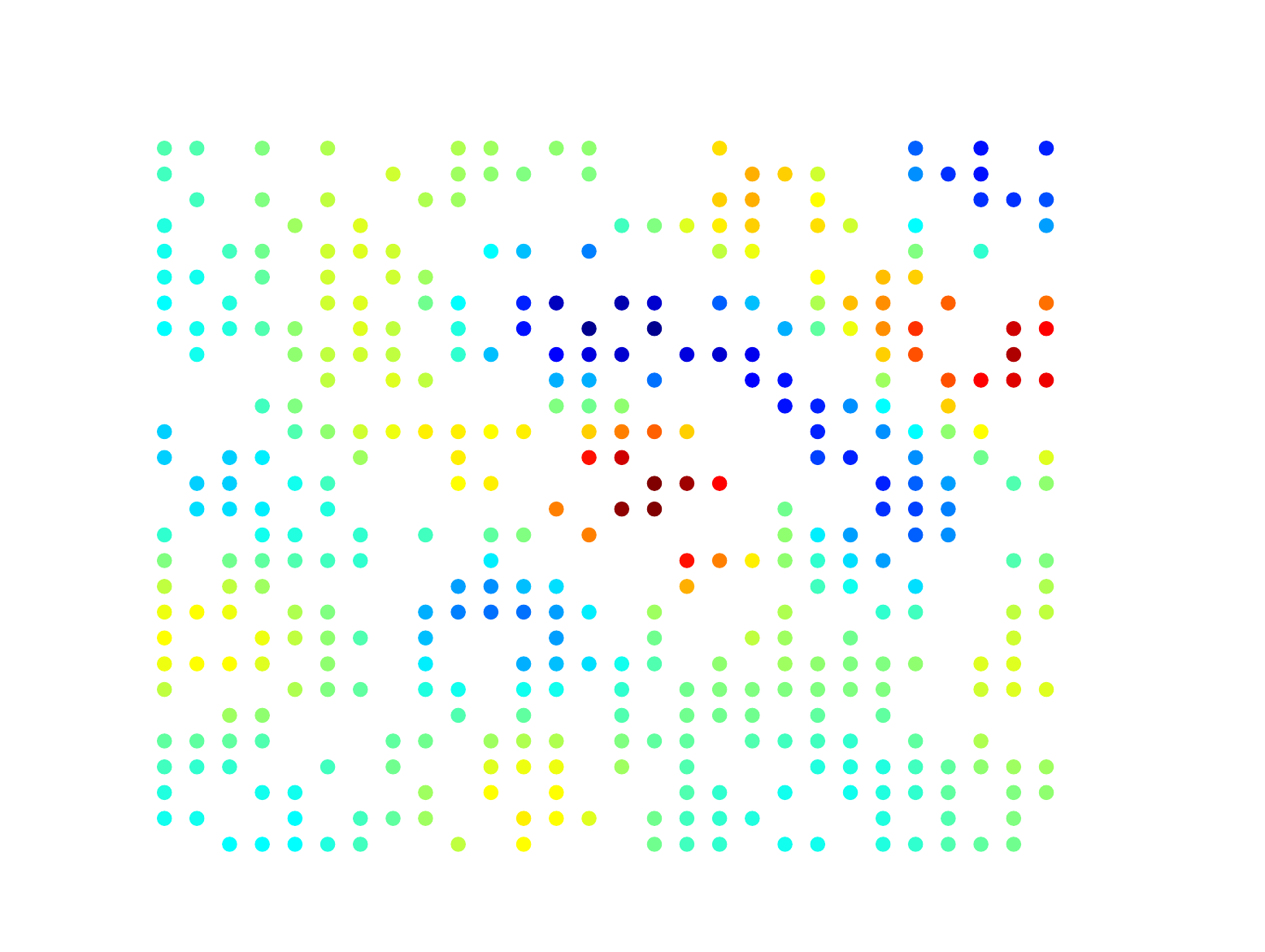}} 
\caption{Examples of Eigenfunctions of the Graph Laplacian $v_2$, $v_{20}$.} 
\label{mnistgrid_freq}
\end{figure}

We first apply the constructions from sections \ref{spectral_sec} and \ref{spatial_sec} 
to the subsampled MNIST dataset.
Figure \ref{mnistgrid_ex} shows examples of the resulting input signals, and Figures
\ref{mnistgrid_space}, \ref{mnistgrid_freq} show the hierarchical clustering constructed
from the graph and some eigenfunctions of the graph Laplacian, respectively. 
The performance of various graph architectures is reported in 
Table \ref{mnistgrid_results}. 
To serve as a baseline, we compute the standard Nearest Neighbor classifier, 
which performs slightly worse than in the full MNIST dataset ($2.8\%$).
A two-layer Fully Connected neural network reduces the error to $1.8\%$. 
The geometrical structure of the data can be exploited with the CNN graph architectures. 
 Local Receptive Fields adapted to the graph structure outperform the fully connected 
network. In particular, two layers of filtering and max-pooling define a network which efficiently
aggregates information to the final classifier. 
The spectral construction performs slightly worse on this dataset. We considered a frequency cutoff 
of $N/2=200$. However, the frequency smoothing architecture described in 
section \ref{smoothspect}, which contains the smallest
number of parameters, outperforms the regular spectral construction.

These results can be interpreted as follows. MNIST digits are characterized
by localized oriented strokes, which require measurements with good spatial 
localization. Locally receptive fields are constructed to explicitly satisfy this constraint, 
whereas in the spectral construction the measurements are not enforced to become
spatially localized. Adding the smoothness constraint on the spectrum of the filters 
improves classification results, since the filters are enforced to have better 
spatial localization. 

\begin{figure}[h]
\centering
\subfigure[ ]{
\includegraphics[scale=0.4]{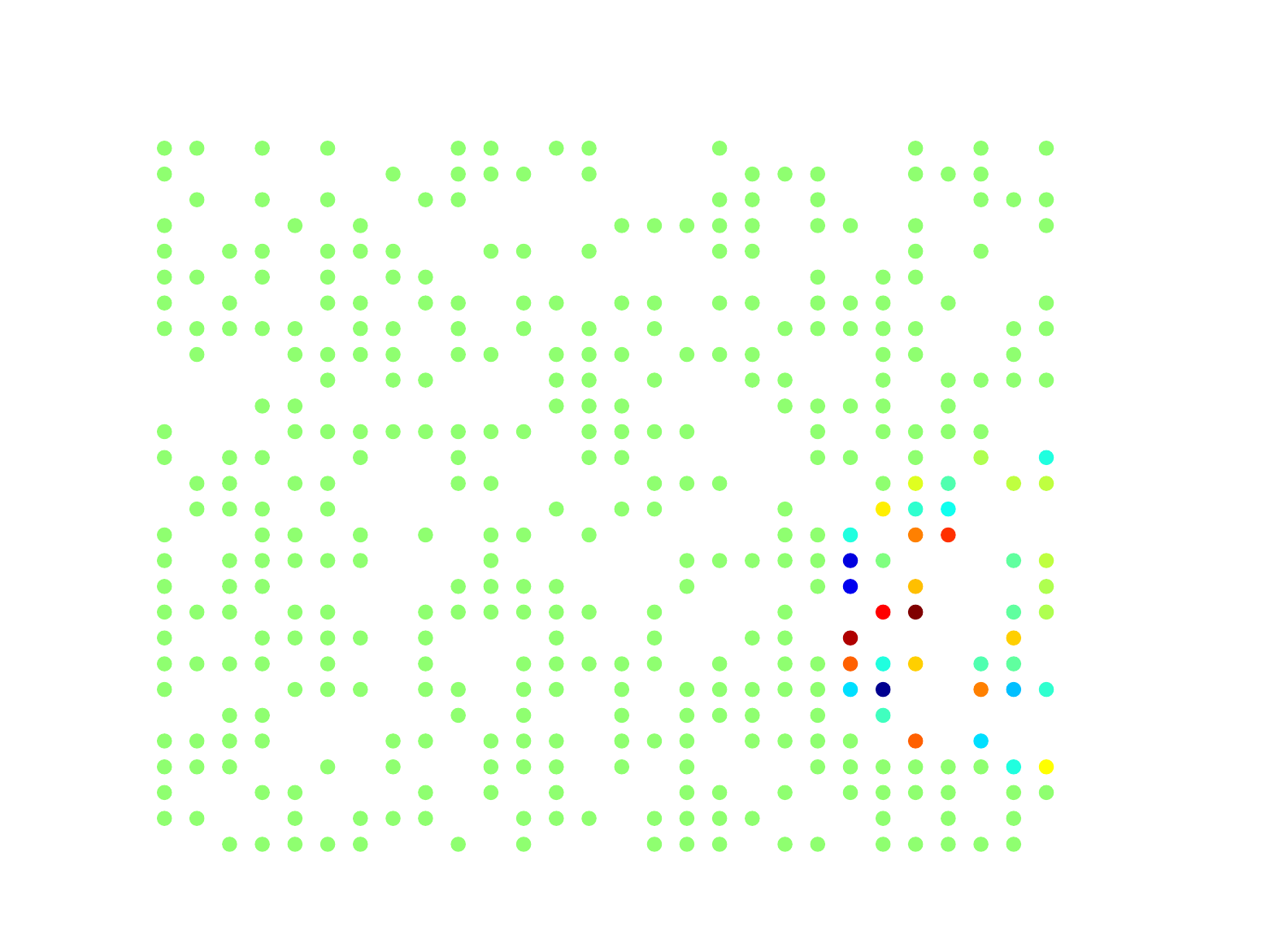}}
\subfigure[ ]{
\includegraphics[scale=0.4]{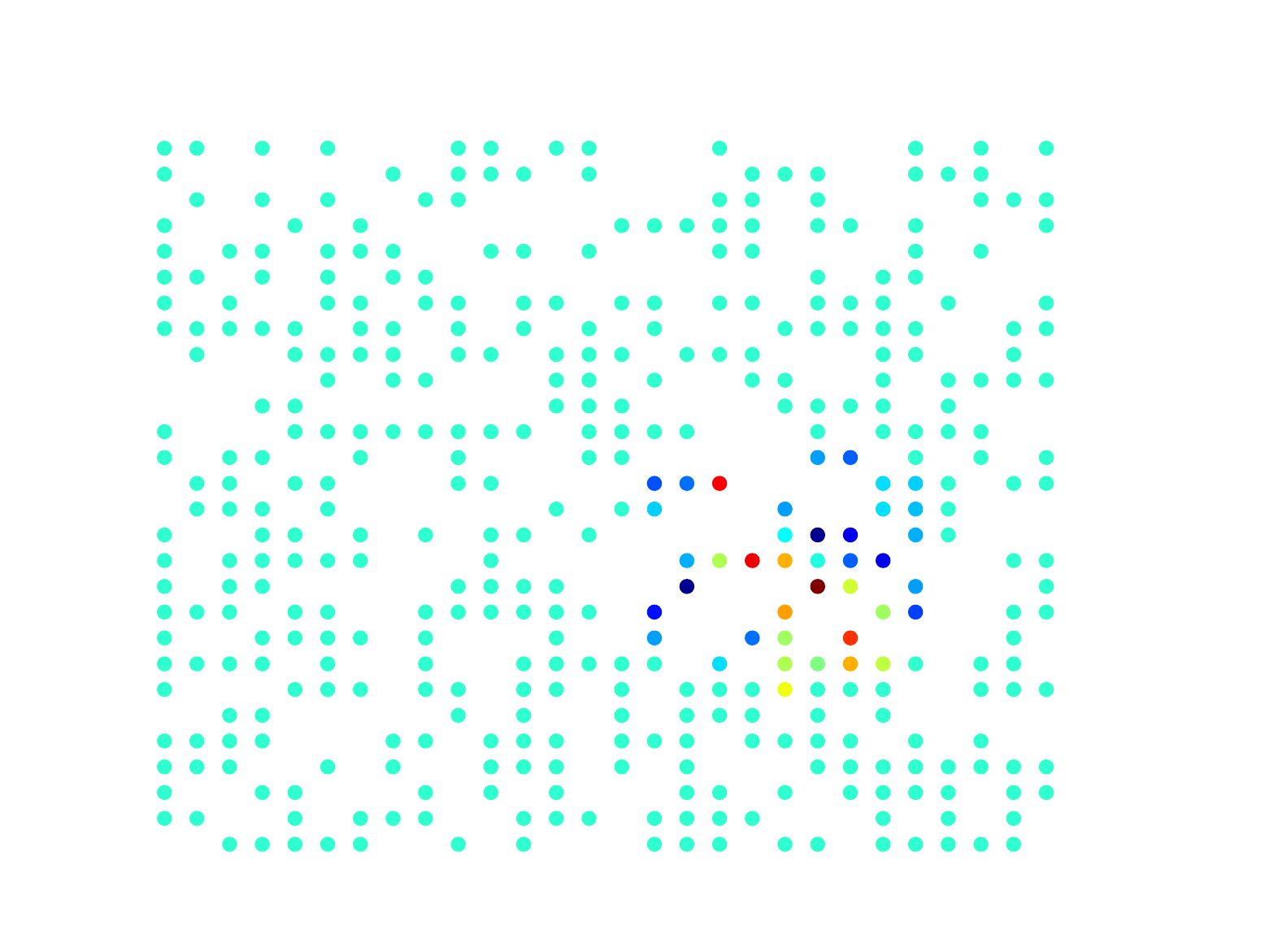}} \\
\subfigure[ ]{
\includegraphics[scale=0.4]{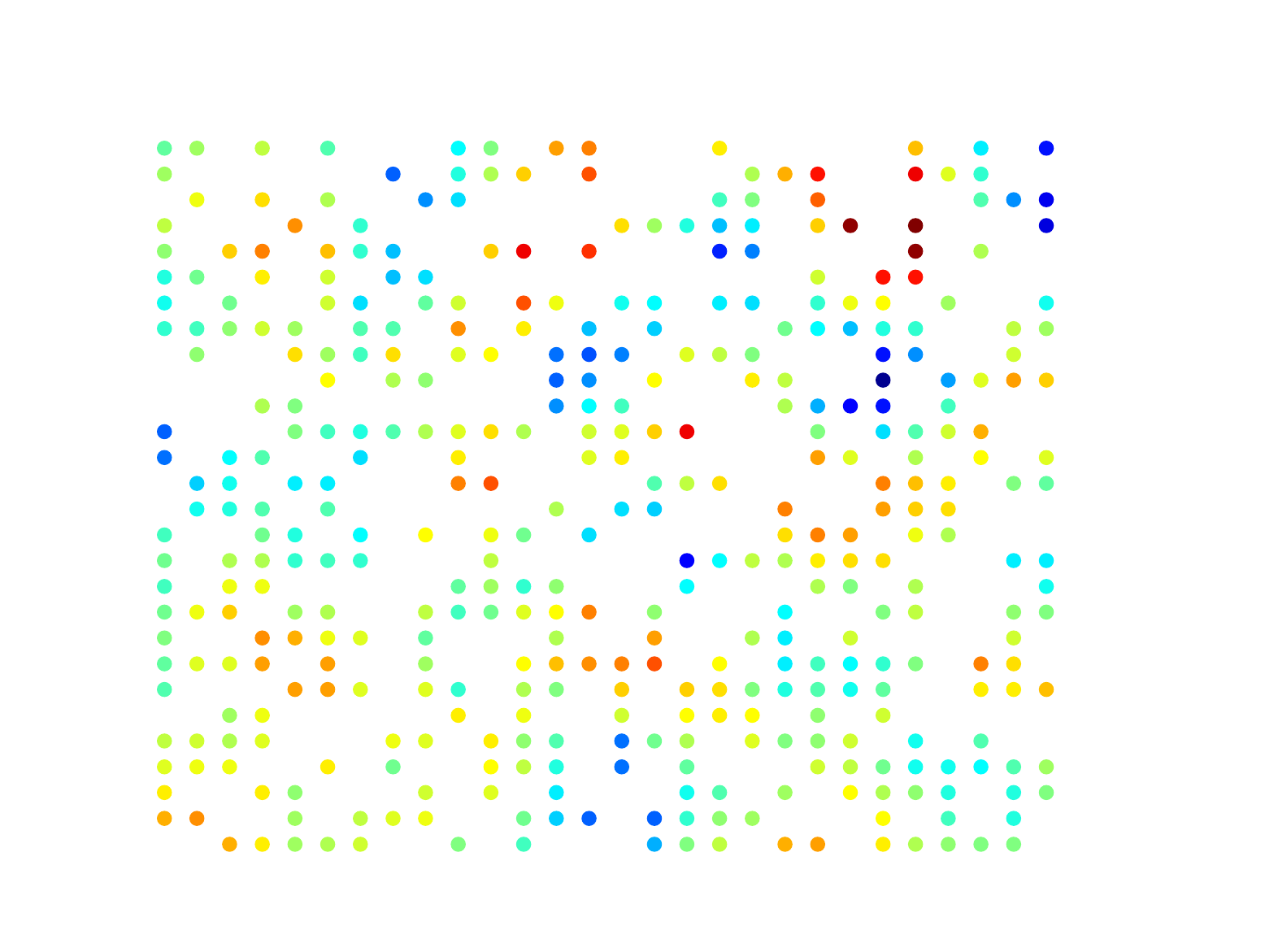}}
\subfigure[ ]{
\includegraphics[scale=0.4]{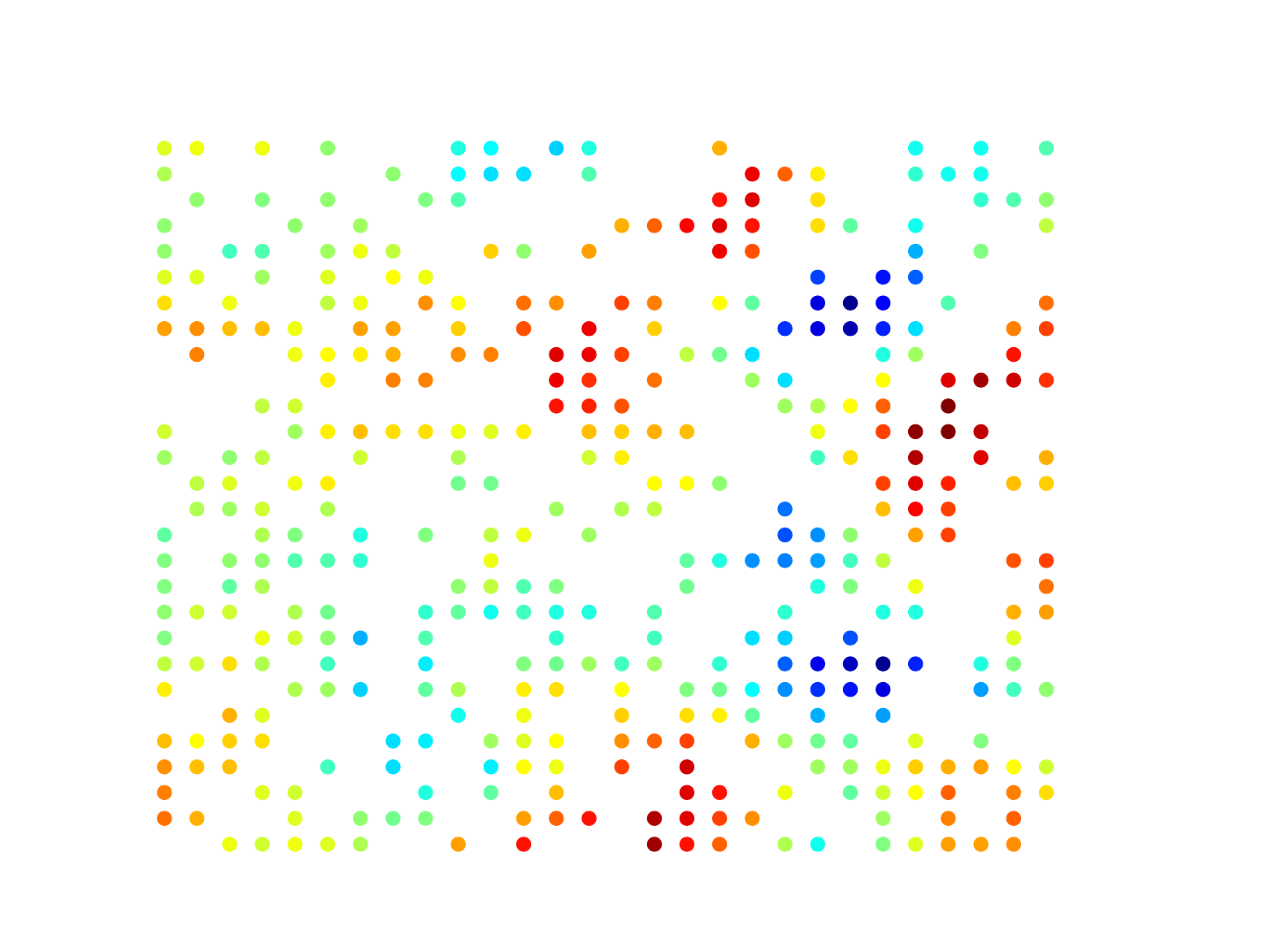}} \\
\subfigure[ ]{
\includegraphics[scale=0.4]{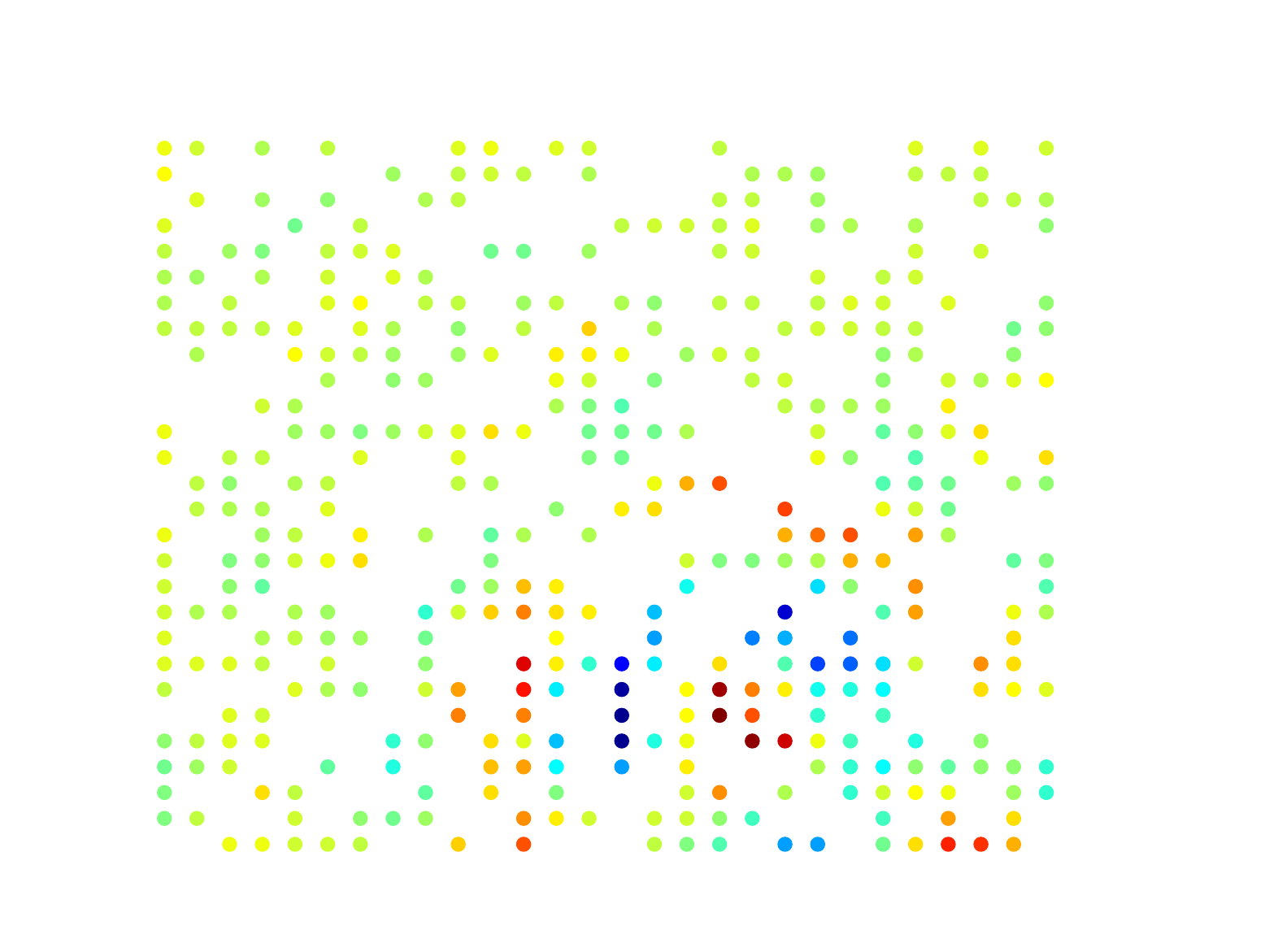}}
\subfigure[ ]{
\includegraphics[scale=0.4]{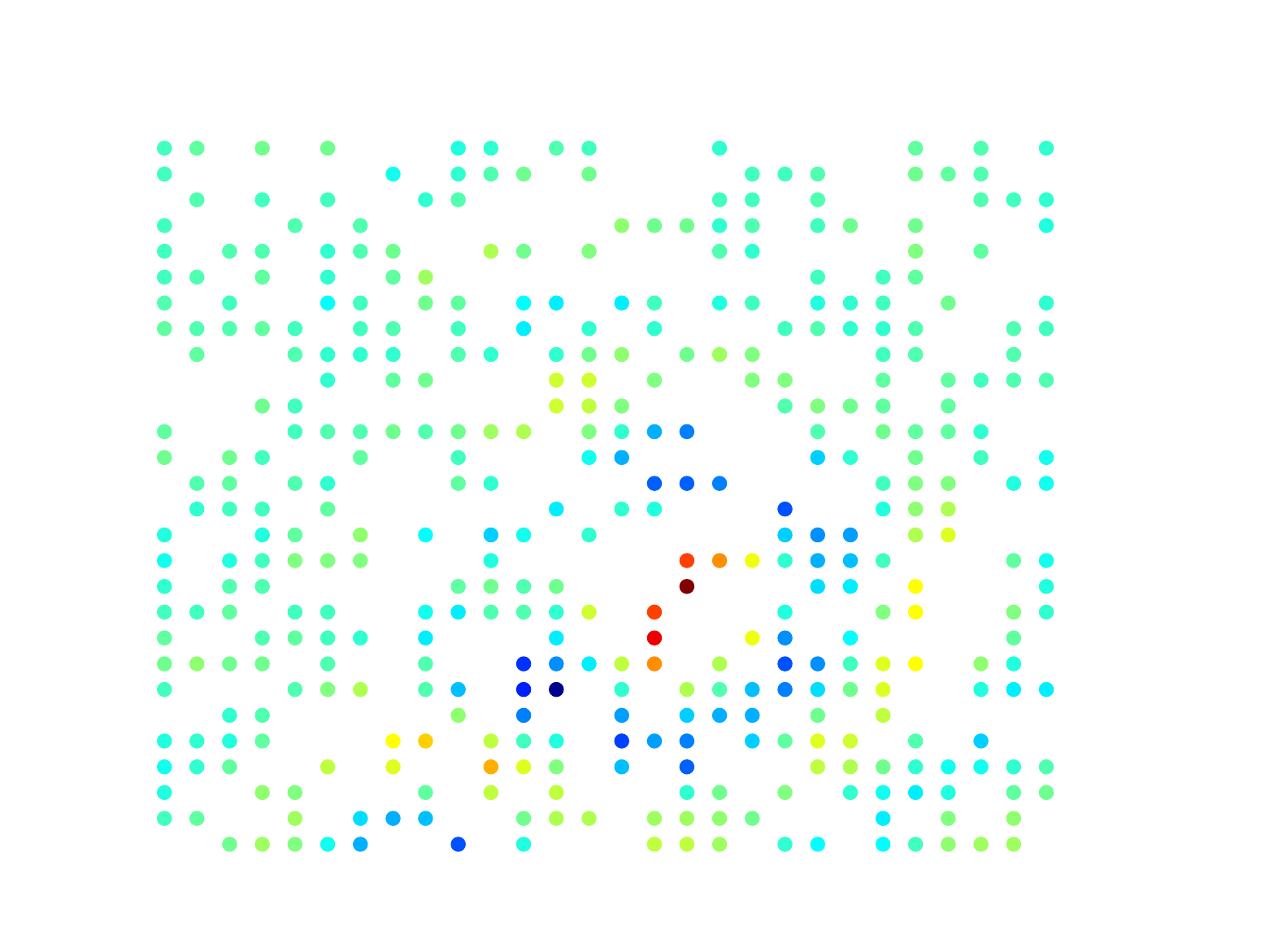}} 
\caption{Subsampled MNIST learnt filters using spatial 
and spectral construction. (a)-(b) Two different receptive fields encoding the same feature
in two different clusters. (c)-(d)  Example of a filter obtained with the spectral construction. 
(e)-(f) Filters obtained with smooth spectral construction.} 
\label{mnistgrid_filters}
\end{figure}

This fact is illustrated in Figure \ref{mnistgrid_filters}. We verify that 
Locally Receptive fields encode different templates across different 
spatial neighborhoods, since there is no global strucutre tying them together. 
On the other hand, spectral constructions have the capacity to generate
local measurements that generalize across the graph. When the spectral
multipliers are not constrained, the resulting filters tend to be spatially 
delocalized, as shown in panels (c)-(d). This corresponds to the 
fundamental limitation of Fourier analysis to encode local phenomena. 
However, we observe in panels (e)-(f) that a simple smoothing across the 
spectrum of the graph Laplacian restores some form of spatial localization 
and creates filters which generalize across different spatial positions, 
as should be expected for convolution operators.

\begin{table}
\caption{Classification results on MNIST
subsampled on $400$ random locations, for different architectures. 
FC$N$ stands for a fully connected layer with $N$ outputs, 
LRF$N$ denotes the locally connected construction from Section \ref{spatial_sec} 
with $N$ outputs, MP$N$ is a max-pooling layer with $N$ outputs, 
and SP$N$ stands for the spectral layer from Section \ref{spectral_sec}.
}
\label{mnistgrid_results}
\begin{center}
\begin{tabular}{|c|c |c |}
\hline
method & Parameters & Error \\
\hline
Nearest Neighbors & N/A & $4.11$ \\
\hline
400-FC800-FC50-10 & $3.6 \cdot 10^{5}$ & $1.8$ \\
\hline
400-LRF1600-MP800-10 & $7.2 \cdot 10^{4} $ & $1.8$ \\
400-LRF3200-MP800-LRF800-MP400-10 & $1.6 \cdot 10^{5} $ & ${\bf 1.3}$ \\
\hline
400-SP1600-10 ($d_1=300$, $q=n$) & $3.2 \cdot 10^{3}$ &  $2.6$ \\
400-SP1600-10 ($d_1=300$, $q=32$) & $1.6 \cdot 10^{3}$  & $2.3$ \\ 
400-SP4800-10 ($d_1=300$, $q=20$) & $5 \cdot 10^{3} $ & $1.8$ \\ 
\hline
\end{tabular}
\end{center}
\end{table}

\subsection{MNIST on the sphere}
\label{sphereMNISTsect}

\begin{figure}[h]
\centering
\subfigure[ ]{
\includegraphics[scale=0.4]{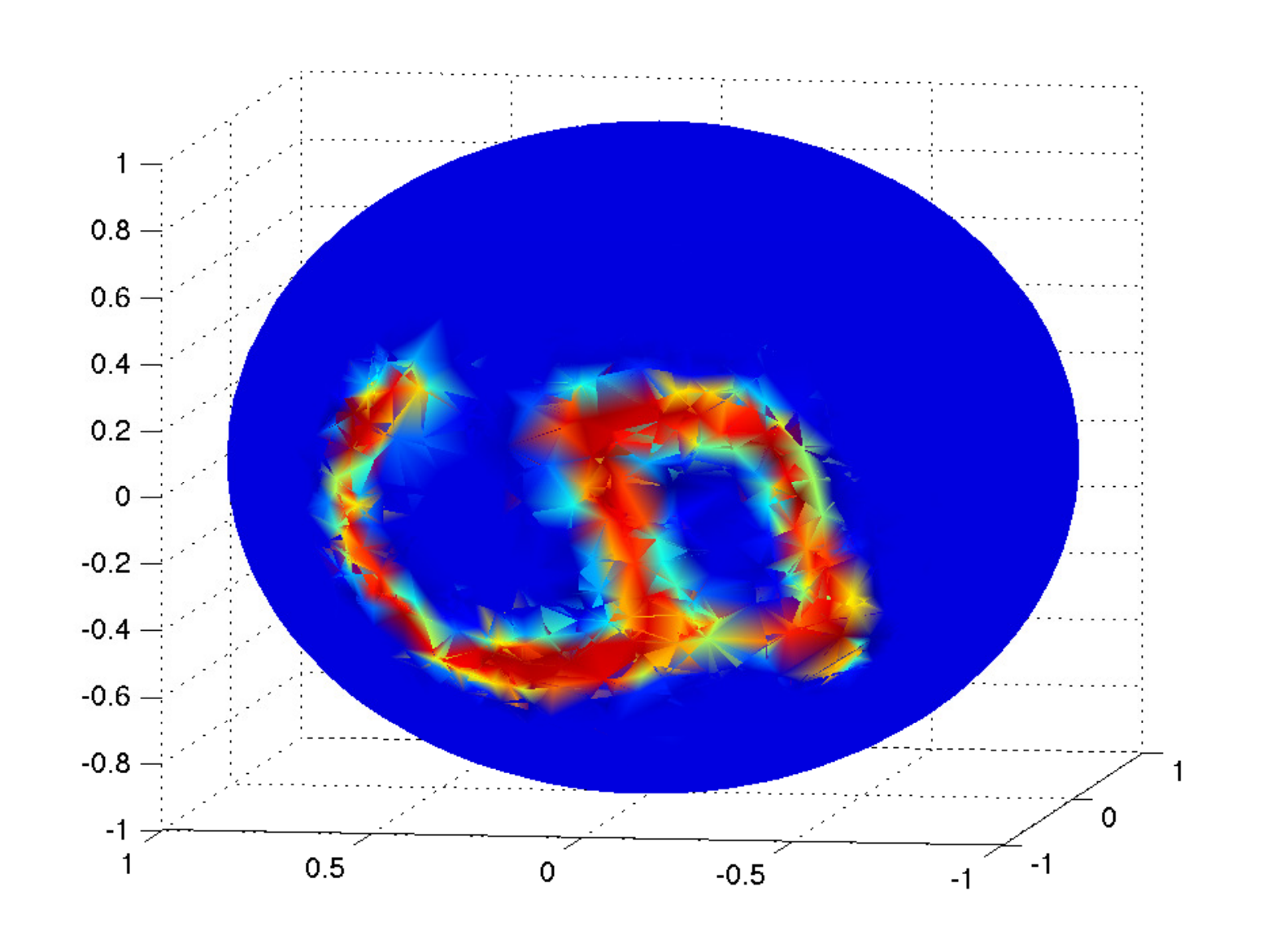}}
\subfigure[ ]{
\includegraphics[scale=0.4]{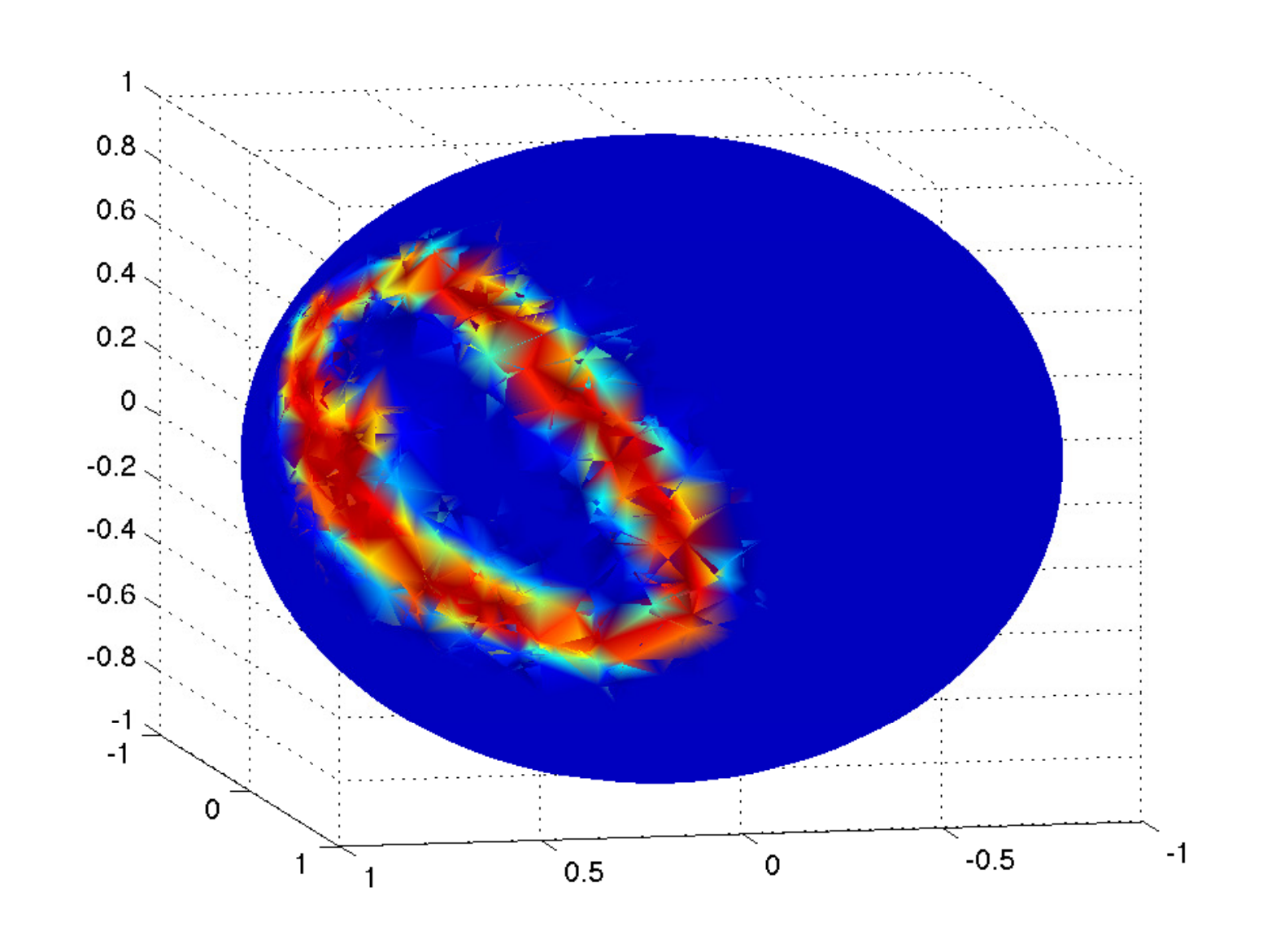}} 
\caption{Examples of some MNIST digits on the sphere.} 
\label{mnistsphere_ex}
\end{figure}

\begin{figure}[h]
\centering
\subfigure[ ]{
\includegraphics[scale=0.4]{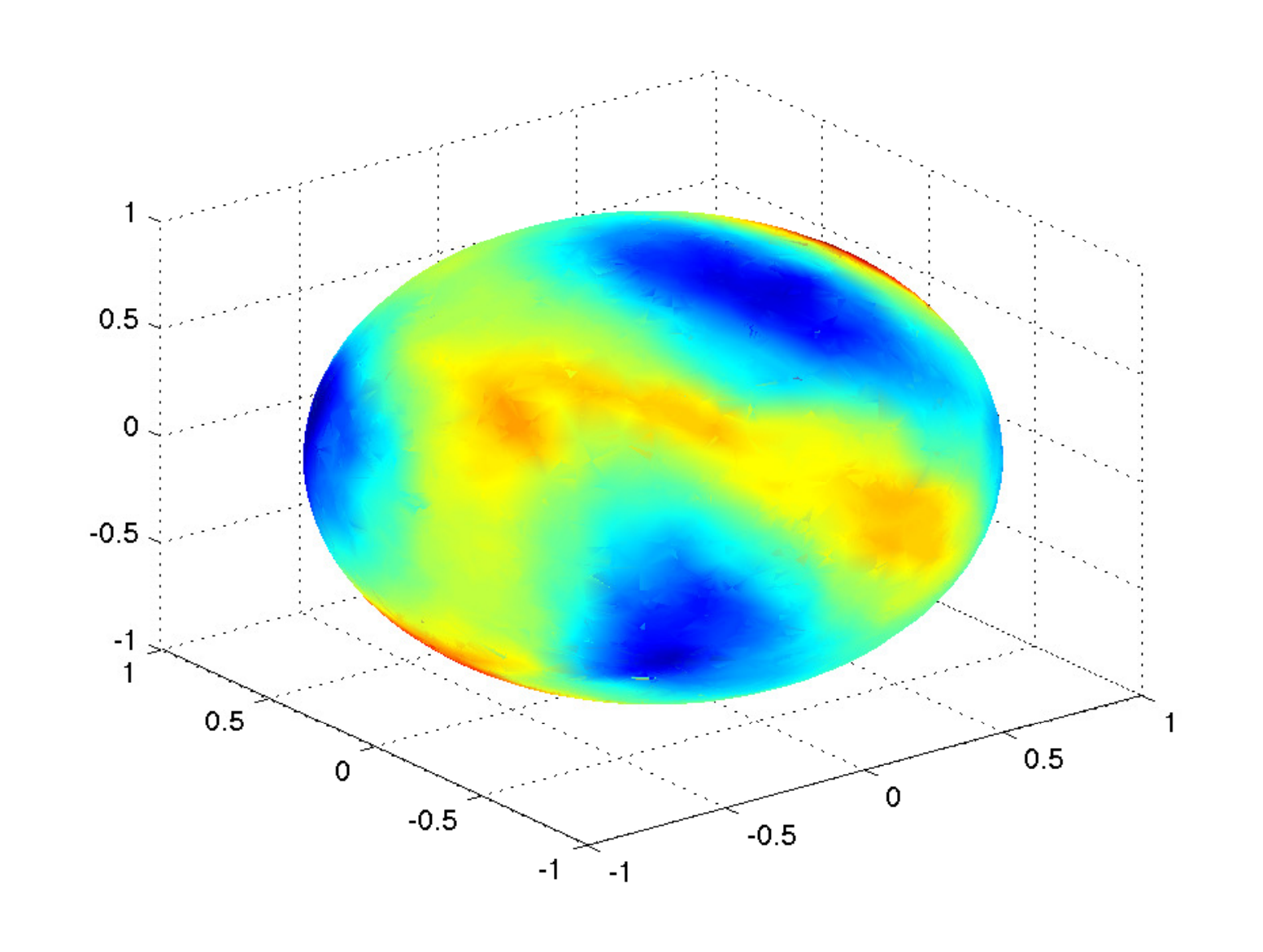}}
\subfigure[ ]{
\includegraphics[scale=0.4]{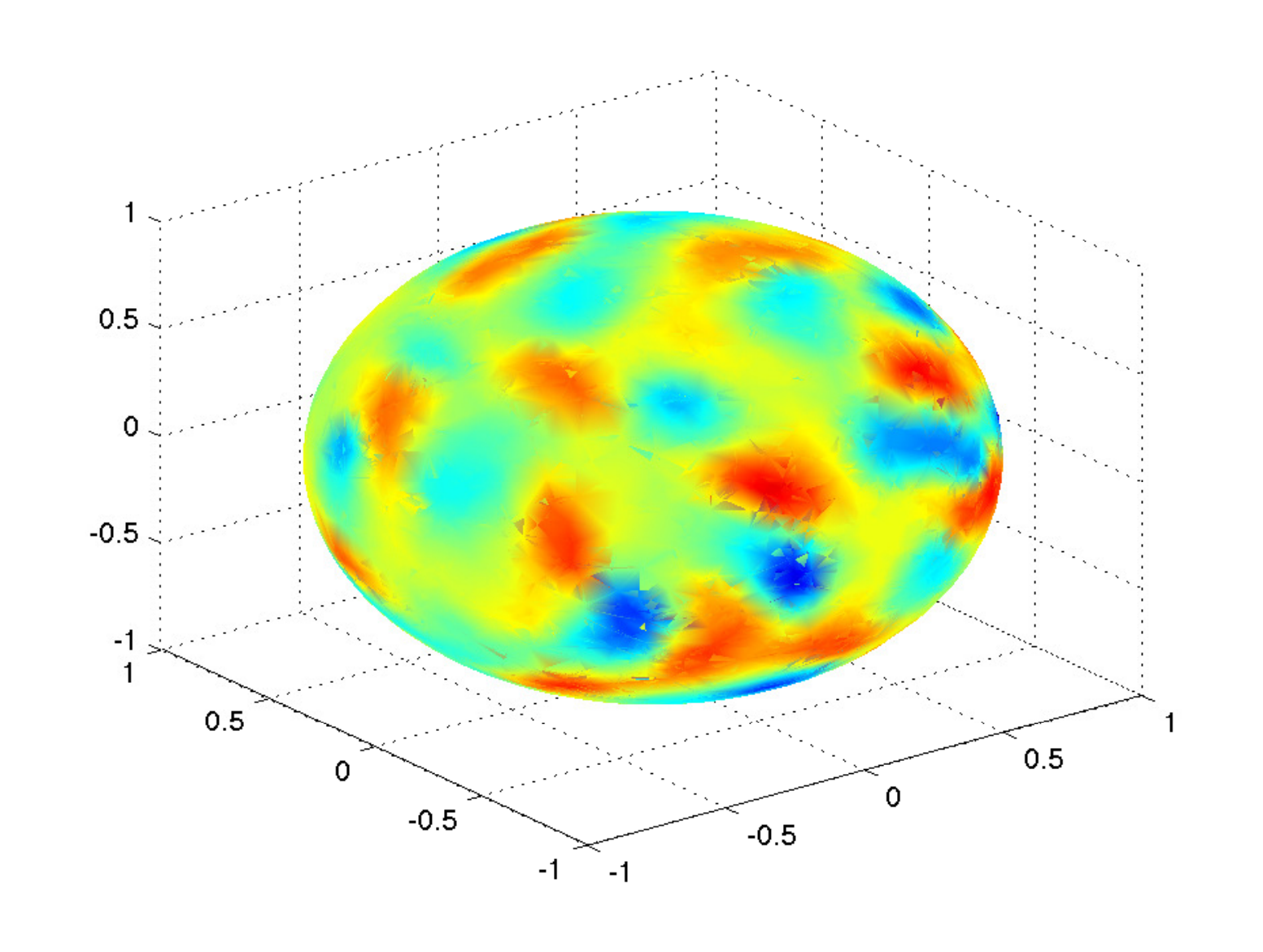}} 
\caption{Examples of Eigenfunctions of the Graph Laplacian $v_{20}$, $v_{100}$} 
\label{mnistsphere_freq}
\end{figure}

We test in this section the graph CNN constructions on another low-dimensional 
graph. In this case, we lift the MNIST digits to the sphere.  
The dataset is constructed as follows.
We first sample $4096$ random points $S=\{s_j\}_{j\leq 4096}$ 
from the unit sphere $S^2 \subset \mathbb{R}^3$.
We then consider an orthogonal basis ${\bf E}=(e_1,e_2,e_3)$ of $\R^3$ 
with $\| e_1 \| = 1\,,\,  \| e_2 \|=2 \,,\, \|e_3 \|=3  $ and 
a random covariance operator $\Sigma = ({\bf E} + W)^T ({\bf E} + W)$, 
where $W$ is a Gaussian iid matrix with variance $\sigma^2 < 1$.
For each signal $x_i$ from the original MNIST dataset, we sample a covariance
operator $\Sigma_i$ from the former distribution and consider its PCA basis $U_i$.
This basis defines a point of view and in-plane rotation which we use to project 
$x_i$ onto $S$ using bicubic interpolation. 
Figure \ref{mnistsphere_ex} shows examples of the
resulting projected digits. Since the digits `6' and `9' are equivalent
modulo rotations, we remove the `9' from the dataset.
Figure \ref{mnistsphere_freq} shows two eigenvectors of the graph 
Laplacian.

\begin{figure}[h]
\centering
\subfigure[ ]{
\includegraphics[scale=0.4]{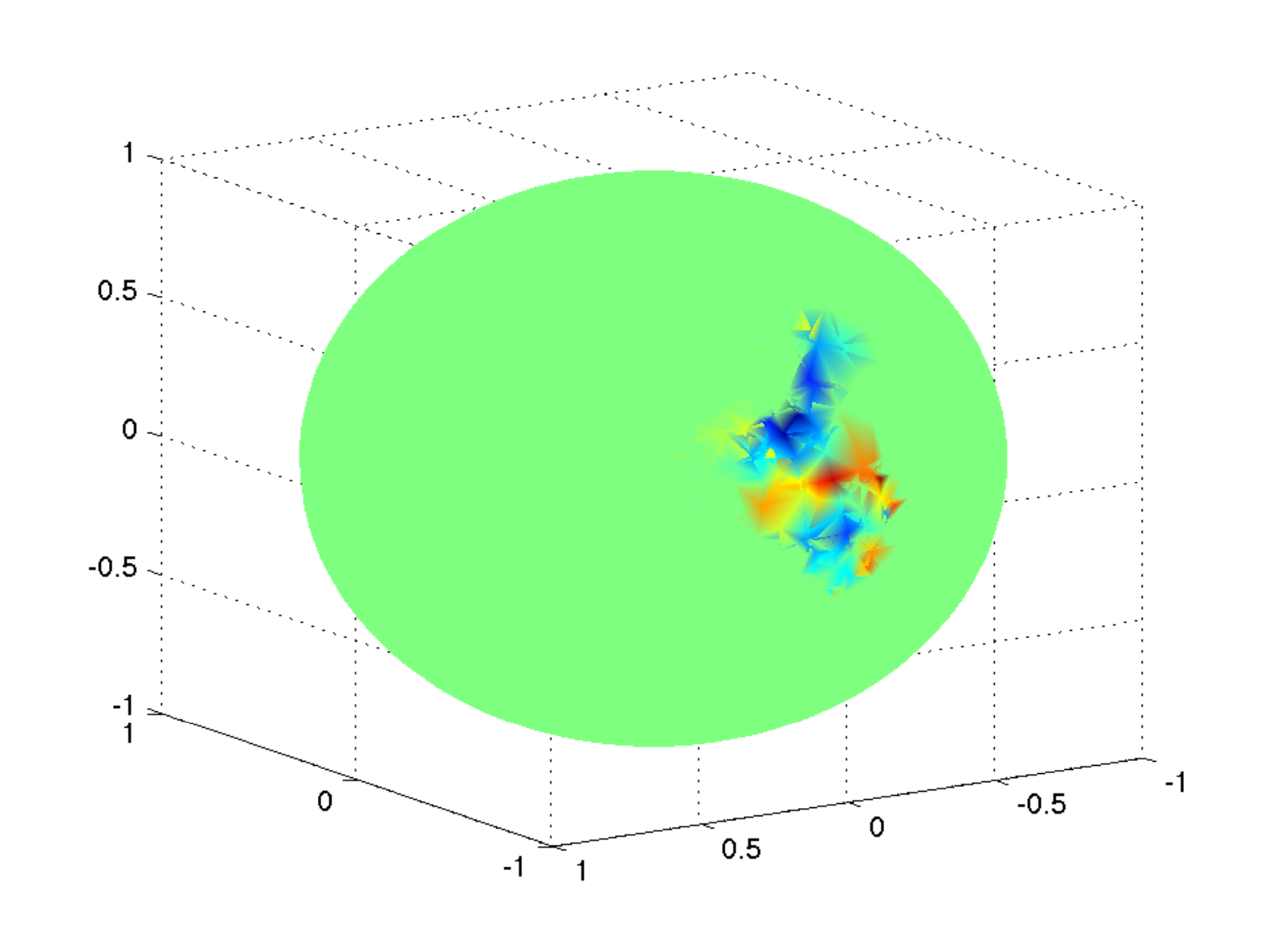}}
\subfigure[ ]{
\includegraphics[scale=0.4]{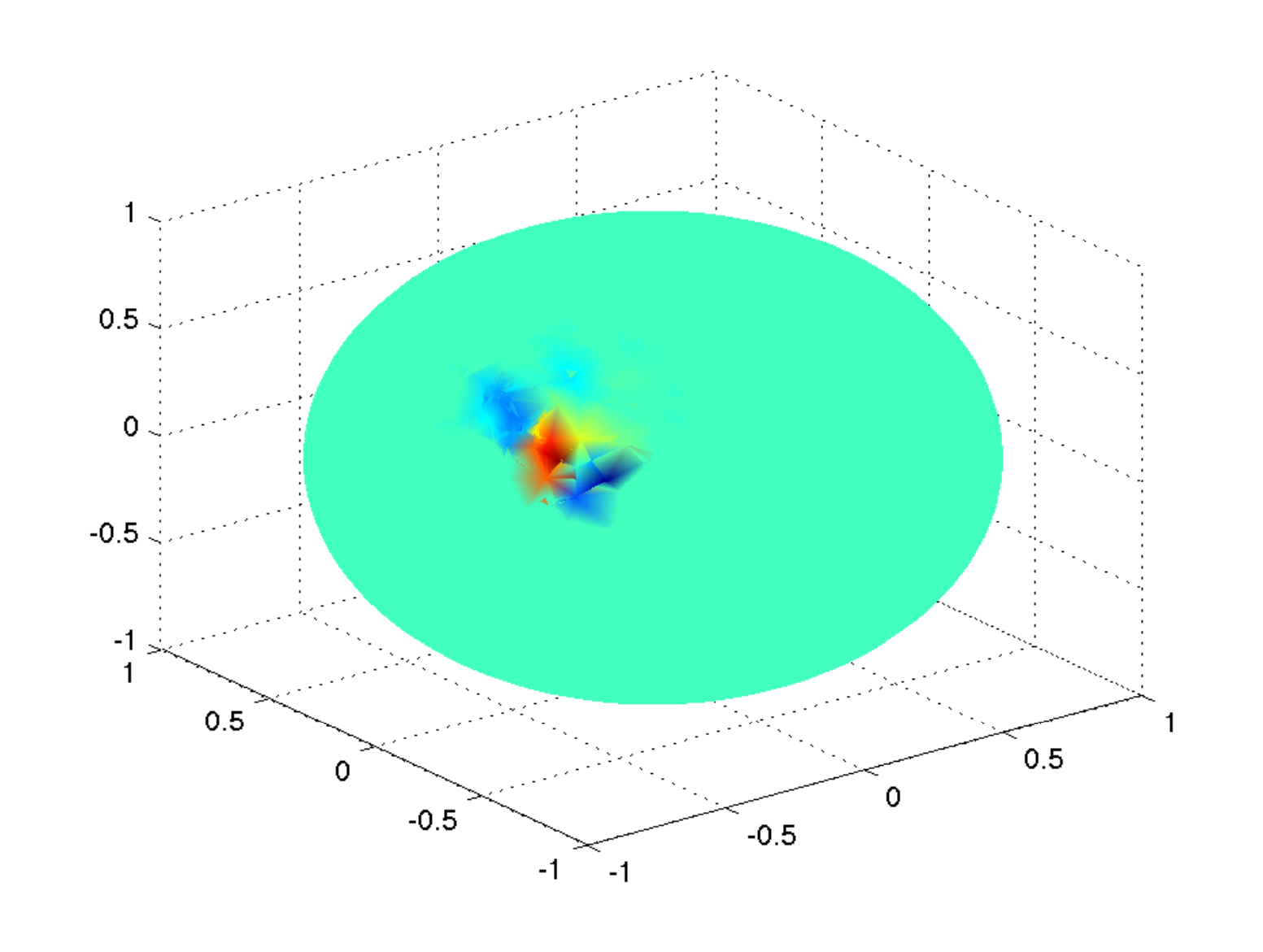}} \\
\subfigure[ ]{
\includegraphics[scale=0.4]{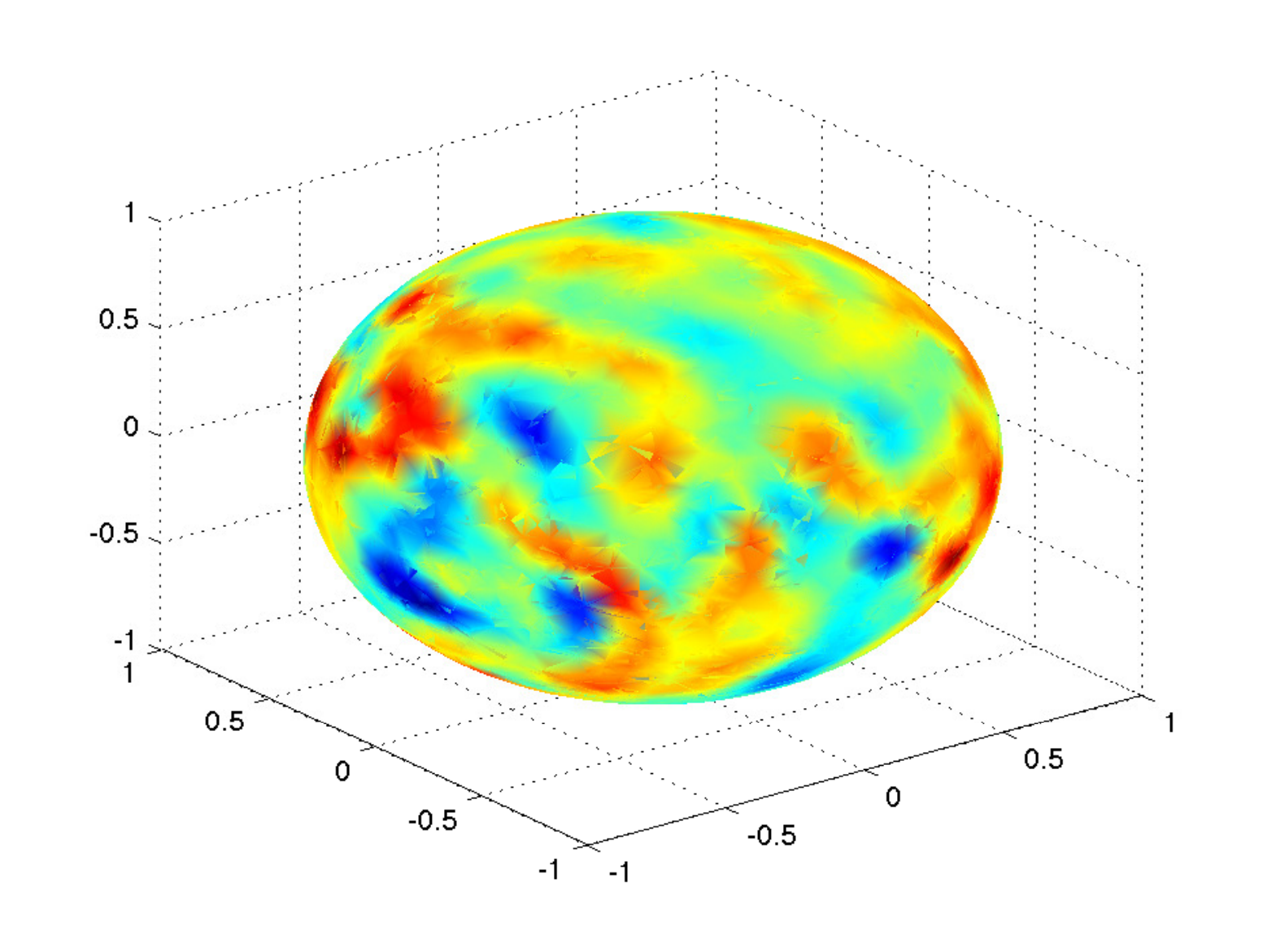}}
\subfigure[ ]{
\includegraphics[scale=0.4]{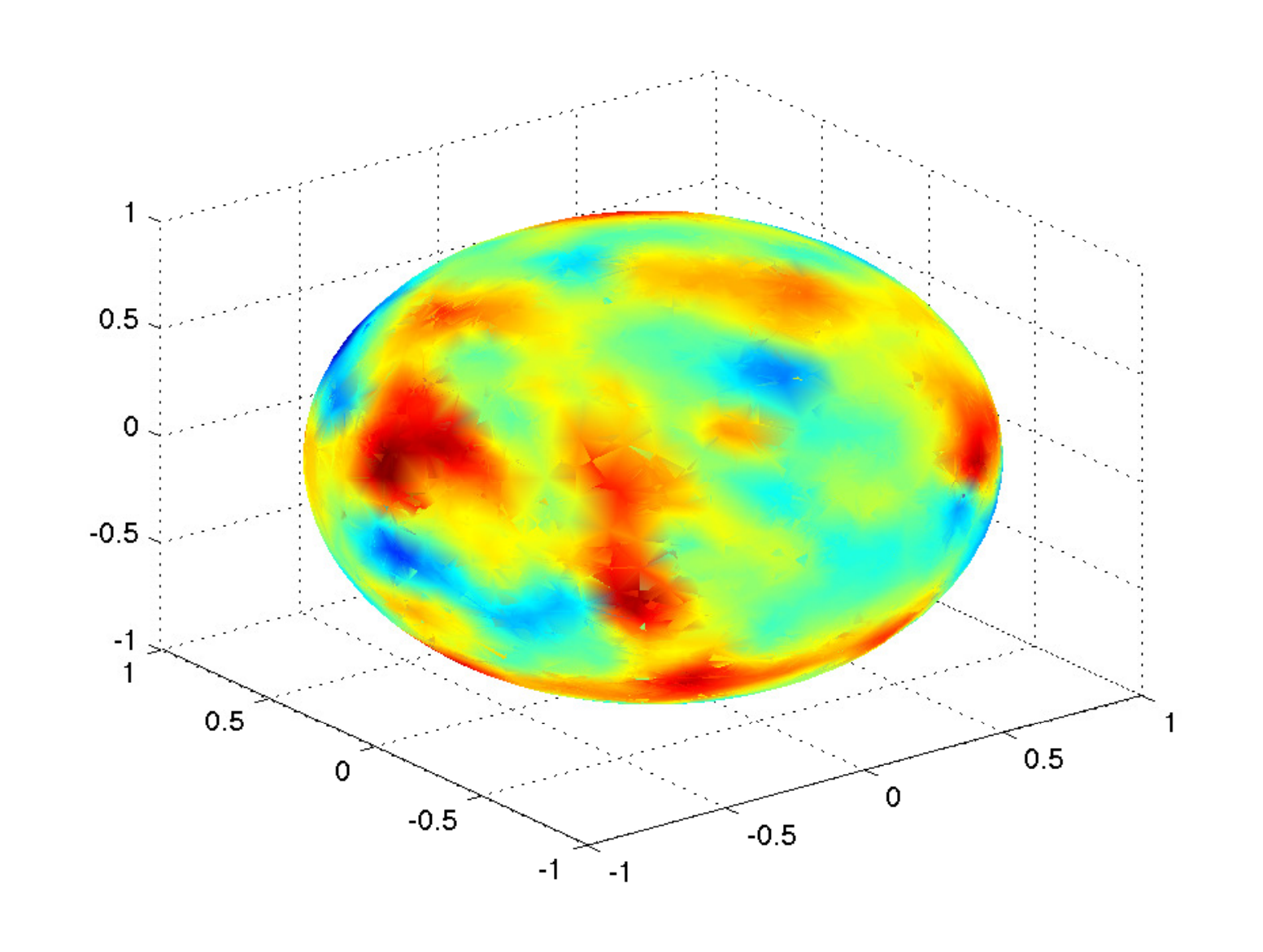}} \\
\subfigure[ ]{
\includegraphics[scale=0.4]{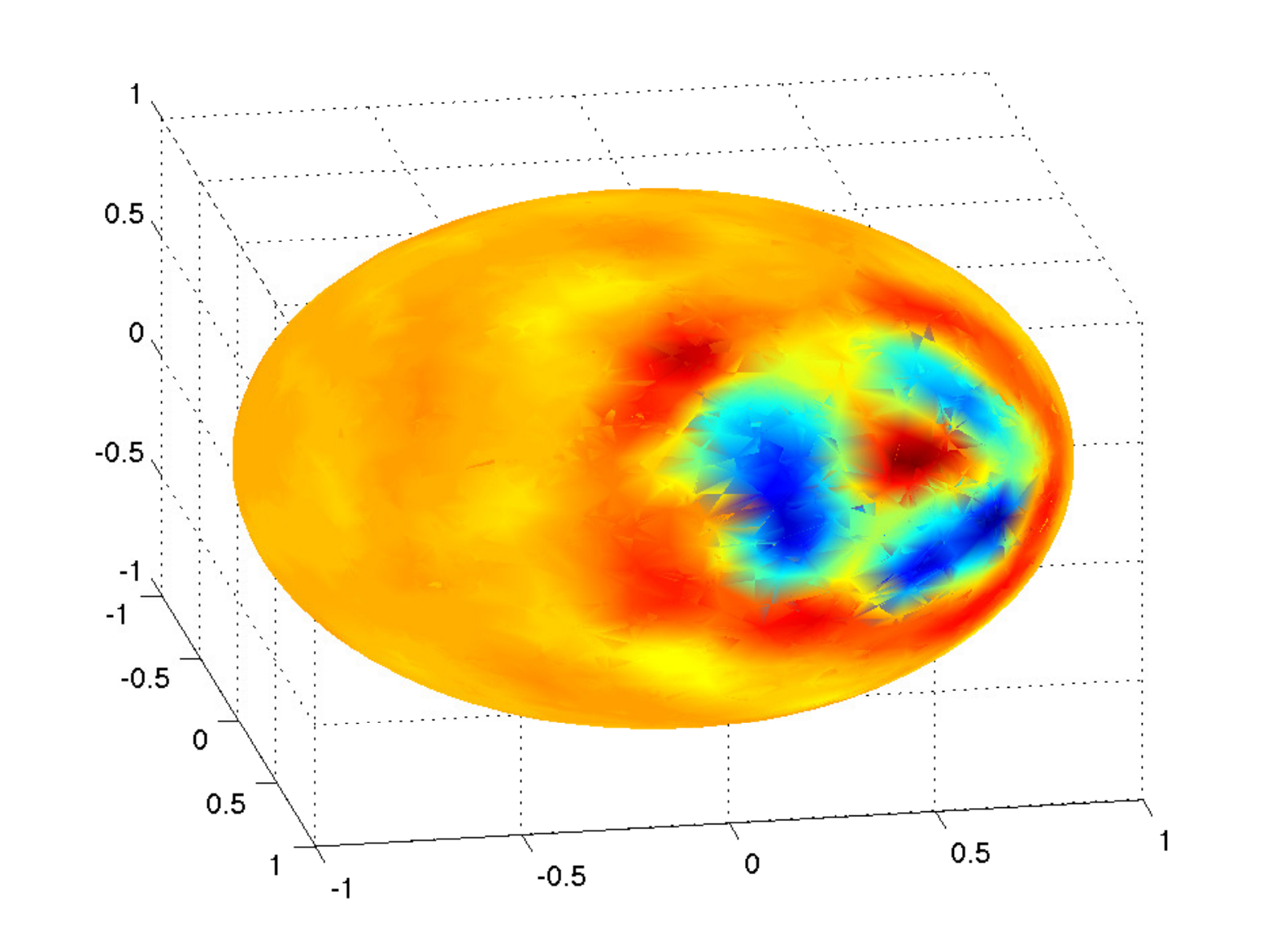}}
\subfigure[ ]{
\includegraphics[scale=0.4]{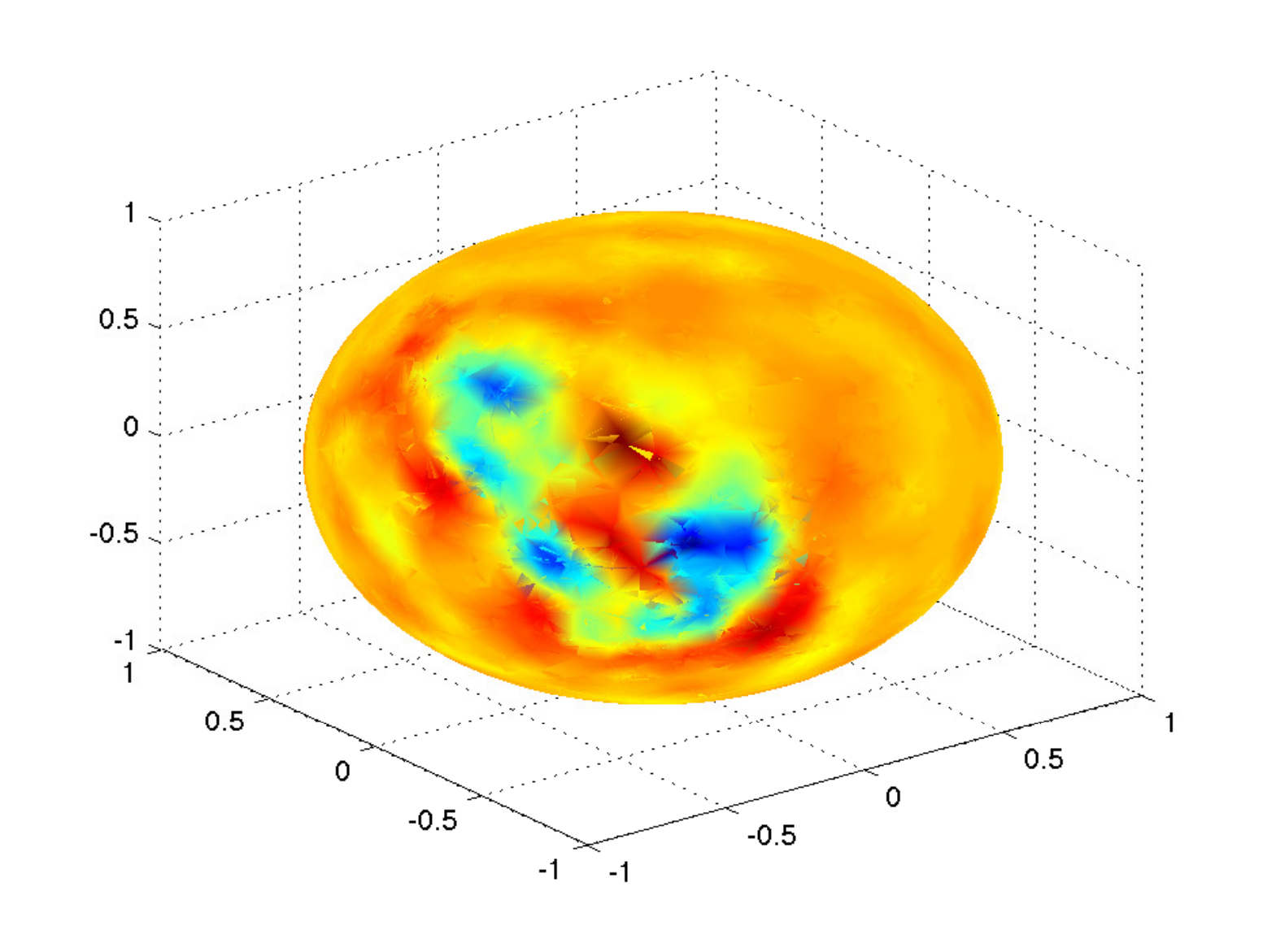}} 
\caption{Filters learnt on the MNIST-sphere dataset, using spatial 
and spectral construction. (a)-(b) Two different receptive fields encoding the same feature
in two different clusters. (c)-(d)  Example of a filter obtained with the spectral construction. 
(e)-(f) Filters obtained with smooth spectral construction.} 
\label{mnistsphere_filters}
\end{figure}

We first consider ``mild" rotations with $\sigma^2=0.2$. The effect of such 
rotations is however not negligible. Indeed, table \ref{mnistsphere_results} shows 
that the Nearest Neighbor classifer performs considerably worse than in 
the previous example. All the neural network architectures we considered 
significatively improve over this basic classifier. Furthermore, we observe
that both convolutional constructions match the fully connected 
constructions with far less parameters (but in this case, do not improve its performance). Figure \ref{mnistsphere_filters} 
displays the filters learnt using different constructions. Again, we verify that the
smooth spectral construction consistently improves the performance, 
and learns spatially localized filters, even using the naive $1$-D organization of eigenvectors, 
which detect similar features across different locations of the graph (panels (e)-(f)). 

Finally, we consider the uniform rotation case, where now the basis $U_i$ is 
a random basis of $\mathbb{R}^3$. In that case, the intra-class variability 
is much more severe, as seen by inspecting the performance of the Nearest neighbor
classifier. All the previously described neural network architectures 
significantly improve over this classifier, although the performance 
is notably worse than in the mild rotation scenario.
In this case, an efficient representation needs to be fully roto-translation 
invariant. Since this is a non-commutative group, it is likely that
deeper architectures perform better than the models considered here.

\begin{table}
\caption{Classification results on the MNIST-sphere dataset 
generated using partial rotations, 
for different architectures}
\label{mnistsphere_results}
\begin{center}
\begin{tabular}{|c|c |c |}
\hline
method & Parameters & Error \\
\hline
Nearest Neighbors & N/A & $19$ \\
\hline
4096-FC2048-FC512-9 & $10^7 $ & ${\bf 5.6}$ \\
\hline
4096-LRF4620-MP2000-FC300-9 & $8 \cdot 10^{5}$ & ${\bf 6}$ \\
4096-LRF4620-MP2000-LRF500-MP250-9 & $2 \cdot 10^{5} $ & $ 6.5$ \\
\hline
4096-SP32K-MP3000-FC300-9 ($d_1=2048$, $q=n$) & $9 \cdot 10^5 $ & $7$ \\
4096-SP32K-MP3000-FC300-9 ($d_1=2048$, $q=64$) & $9 \cdot 10^5 $ & ${\bf 6}$ \\
\hline
\end{tabular}
\end{center}
\end{table}

\begin{table}
\caption{Classification results on the MNIST-sphere dataset 
generated using uniformly random rotations, 
for different architectures}
\label{mnistsphereh_results}
\begin{center}
\begin{tabular}{|c|c |c |}
\hline
method & Parameters & Error \\
\hline
Nearest Neighbors & NA & $80$ \\
\hline
4096-FC2048-FC512-9 & $10^7 $  & ${ 52}$ \\
\hline
4096-LRF4620-MP2000-FC300-9 & $8 \cdot 10^{5}$ & $61$ \\
4096-LRF4620-MP2000-LRF500-MP250-9 & $2 \cdot 10^{5} $ & $63$ \\
\hline
4096-SP32K-MP3000-FC300-9 ($d_1=2048$, $q=n$) & $9 \cdot 10^5 $  & $56$ \\
4096-SP32K-MP3000-FC300-9 ($d_1=2048$, $q=64$) & $9 \cdot 10^5 $& ${\bf 50}$ \\
\hline
\end{tabular}
\end{center}
\end{table}

\section{Conclusion}
Using graph-based analogues of convolutional architectures can greatly reduce the number of parameters in a neural 
network without worsening (and often improving) the test error, while simultaneously giving a faster forward propagation.  
These methods can be scaled to data with a large number of coordinates that have a notion of locality.

There is much to be done here.  We suspect with more careful training and deeper networks we can consistently 
improve on fully connected networks on ``manifold like'' graphs like the sampled sphere.  
Furthermore, we intend to apply these techniques to less artifical problems, for example, on netflix like 
recommendation problems where there is a biclustering of the data and coordinates.  Finally, the fact that 
smoothness on the naive ordering of the eigenvectors leads to improved results and localized filters suggests 
that it may be possible to make ``dual'' constructions with $O(1)$ parameters per filter in much more generality than the grid.

\bibliographystyle{plain}
\bibliography{refs}

\end{document}